\documentclass{article}

\usepackage[final]{neurips_2021}

\usepackage[utf8]{inputenc} 
\usepackage[T1]{fontenc}    
\usepackage{hyperref}       
\usepackage{url}            
\usepackage{booktabs}       
\usepackage{amsfonts}       
\usepackage{nicefrac}       
\usepackage{microtype}      
\usepackage{xcolor}         
\usepackage{graphicx}
\usepackage{amsmath}
\usepackage{amsthm}
\usepackage[export]{adjustbox}
\usepackage{wrapfig}
\usepackage{placeins}

\newtheorem{prop}{Proposition}

\title{A Critical Look at the Consistency of Causal Estimation with Deep Latent Variable Models}

\author{%
  Severi Rissanen \\
  Department of Computer Science\\
  Aalto University\\
  Espoo, Finland \\
  \texttt{severi.rissanen@aalto.fi}
  \And
  Pekka Marttinen \\
  Department of Computer Science\\
  Aalto University\\
  Espoo, Finland \\
  \texttt{pekka.marttinen@aalto.fi}
}

\begin{document}

\maketitle

\begin{abstract}
  Using deep latent variable models in causal inference has attracted considerable interest recently, but an essential open question is their ability to yield consistent causal estimates. While they have demonstrated promising results and theory exists on some simple model formulations, we also know that causal effects are not even identifiable in general with latent variables. We investigate this gap between theory and empirical results with analytical considerations and extensive experiments under multiple synthetic and real-world data sets, using the causal effect variational autoencoder (CEVAE) as a case study. While CEVAE seems to work reliably under some simple scenarios, it does not estimate the causal effect correctly with a misspecified latent variable or a complex data distribution, as opposed to its original motivation. Hence, our results show that more attention should be paid to ensuring the correctness of causal estimates with deep latent variable models.
\end{abstract}

\section{Introduction}

\begin{wrapfigure}[13]{r}{0.3\linewidth}
    \centering
    \includegraphics[width=1\linewidth]{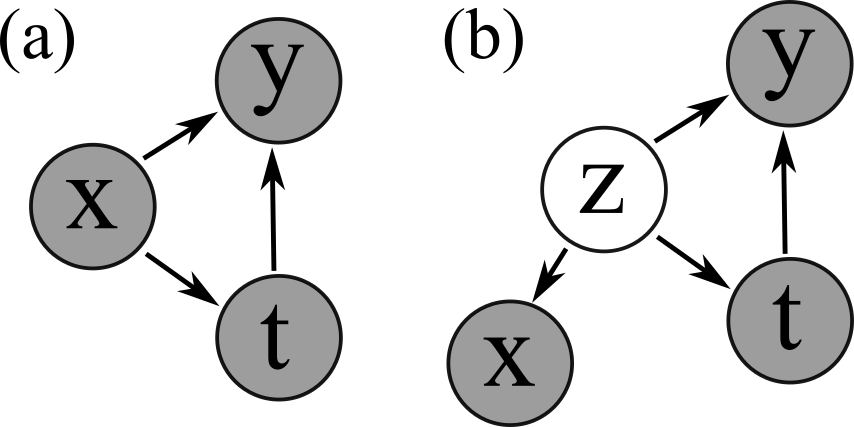}
    \caption{(a) The direct confounding causal graph. (b) The causal graph with an unobserved confounder $z$, and proxy variables $x$. }
    \label{fig:causal_graphs}   
\end{wrapfigure}

Causal inference, dealing with the questions of when and how we can make causal statements based on observational data, has been a topic of growing interest in the deep learning community recently. On the one hand, causal inference promises to provide traditional machine learning and AI with methods for explainability, domain adaptation, and causal reasoning capabilities in general~\citep{pearl2019seven}. On the other hand, many deep learning methods for improving causal inference have been proposed. Some of the models have been designed under the assumption of no unobserved confounding~\citep{shalit2017estimating, yoon2018ganite,shi2019adapting}, while others utilize latent variables in one way or the other to account for unobserved confounders~\citep{louizos2017causal, rakesh2018linked, pfohl2019counterfactual, madras2019fairness, mayer2020missdeepcausal, chen2020inferring, pawlowski2020deep, jesson2020identifying}. Although some simple models with unobserved confounders are known to produce correct results~\citep{angrist1996identification, pearl2016causal, miao2018identifying}, the \emph{consistency}, i.e., whether in the limit of large data the correct causal effect is retrieved, is usually left largely open with deep latent variable models.  


In particular, \citet{louizos2017causal} proposed the causal effect variational autoencoder (CEVAE) for performing causal inference in the setting where we have an unobserved confounder, of which only noisy proxy variables are available. The causal graph is shown in Fig.\ref{fig:causal_graphs}b, which contrasts to the standard "no unobserved confounding", or direct confounding, graph in Fig.\ref{fig:causal_graphs}a. The proxy variable setting has been studied rigorously elsewhere by \citep{pearl2016causal}, where the authors provided provably correct methods for a few simple types of data with strong assumptions about the data generating process. They dubbed the process of estimating causal effects in this context as "effect restoration". CEVAE was proposed to relax these assumptions significantly, but the question of consistency was left unanswered.

To provide insight into the behavior of deep latent variable models as causal effect estimators, we use CEVAE as a case study. CEVAE is a natural choice because it is based on the well-established standard variational autoencoder~\citep{Kingma2014,rezende2014stochastic}, and it allows comparison with analytical, provably correct methods. We conduct rigorous experiments with the model using various synthetic and semi-synthetic data sets and also provide theoretical statements in some special cases. Even though generic theoretical results are difficult to get for variational autoencoders, we provide intuitive conclusions on the assumptions under which the model works or does not work.

\section{Preliminaries}

\subsection{Model description}

The objective in the scenario in Fig.\ref{fig:causal_graphs}b is to learn the causal effect between the variables $t$ and $y$. $t$ is a variable on which we can perform an intervention, e.g., a treatment on a disease. $x$ is the possibly multidimensional proxy providing indirect and noisy information about the unobserved confounder $z$. The correct interventional distribution $p(y|do(t))$ is defined with the formula
\begin{equation}
    p(y|do(t)) = \int p(y|z,t)p(z) dz,\label{eq:truepydot}
\end{equation}
where the integral is replaced with a sum for a discrete confounder. Note that we denote the latent variable of a VAE with $z$ as well, even though they are conceptually separate and might not be distributed in the same way. The distinction should be clear from the context. 

At its core, the causal effect variational autoencoder is simply a regular variational autoencoder with two additional assumptions. First, we assume that the latent variable $z$ corresponds to the unobserved confounder in some way so that after training, we can get the causal effect $p(y|do(t))$ by estimating with the adjustment formula:
\begin{equation}
    p(y|do(t)) \approx p_\theta(y|do(t)) = \int p_\theta(y|z,t) p(z) dz,\label{eq:pydot}
\end{equation}
where $\theta$ are learned decoder parameters and $p(z)$ is the VAE prior. Second, we assert some additional restrictions on the structure of the decoder. The idea is that the quantities reconstructed during training should follow the conditional independencies specified by the causal graph in Fig.\ref{fig:causal_graphs}b so that the conditional probability of observed variables given the latent variable factorizes as
\begin{equation}
    p_\theta(x^{i},t^{i},y^{i}|z) = p_\theta(x^{i}|z)p_\theta(t^{i}|z)p_\theta(y^{i}|z,t^{i}),\label{eq:decoder_factorization}
\end{equation}
where superscript $i$ refers to the $i$:th observation. Thus, we can write the ELBO for the model as
\begin{align}
    \mathcal{L}(\theta,\phi) = \sum_i\big[&\mathbb{E}_{q_\phi(z|x^{i},t^{i},y^{i})}[\log p_\theta(x^{i}|z) + \log p_\theta(t^{i}|z) \nonumber\\
    &+\log p_\theta(y^{i}|z,t^{i})] - \textrm{KL}[q_\phi(z|x^{i},t^{i},y^{i})||p(z)] \big],\label{eq:ELBO}
\end{align}
where $x^{i},t^{i}$ and $y^{i}$ are observed quantities. Thus, we have to define at least four neural networks: The encoder network $q_\phi(z|x^{i},t^{i},y^{i})$ and three decoder networks corresponding to the conditionals in Eq.\ref{eq:decoder_factorization}. The original paper also suggested composing the encoder of multiple networks chosen based on the value of the treatment, but that is not strictly necessary and was not motivated by the causal graph. Note that the decoder differs from usual VAE decoders in that the observed treatment value $t^i$ has to be given as input to the network corresponding to $p_\theta(y|z,t)$. The original paper seemed to suggest that the input should be sampled from the $p_\theta(t|z)$ distribution during training (Fig.2b in \citep{louizos2017causal}), but the factorization in Eq.\ref{eq:decoder_factorization} suggests that the observed $t$ should be used instead.


In terms of the problem statement, CEVAE is perhaps most closely connected to the methods of effect restoration first proposed by \citet{kuroki2014measurement}. They offered provably correct, analytical solutions to the proxy variable problem in Fig.\ref{fig:causal_graphs}b when all variables are jointly normal or categorical and $x$ consists of two variables conditionally independent of each other given $z$. The setting was further studied in~\citep{miao2018identifying}. 
Note that since CEVAE assumes a different causal graph from the direct confounding graph in Fig.\ref{fig:causal_graphs}a, it can't be assumed to give correct results if the true data generating process is the direct confounding graph. For example, \citet{kuroki2014measurement} showed that an observed $p(x,t,y)$ can map to two completely different causal effects by assuming either the direct or the unobserved confounding graphs. As the original CEVAE paper experimented on data with no unobserved confounding in addition to data sets with the unobserved confounding, we suspect that this may have been a point of confusion for some.

\subsection{Possible issues with estimation}\label{sec:estimability}


Theorem 1 in \citep{louizos2017causal} states that if CEVAE is able to recover the $p(z,x,t,y)$ distribution, it is guaranteed to yield correct causal estimates. However, this leaves open the relevant question about when such an assumption can hold, as recovering $p(z,x,t,y)$ entirely is, strictly speaking, impossible due to the unidentifiability of VAEs (e.g. \citep{locatello2019challenging}). In contrast, we don't consider the latent variable of CEVAE to strictly correspond to the true confounder. Instead we view the process of training CEVAE and applying the adjustment formula in Equation \ref{eq:pydot} simply as a statistical estimator for the causal effect, regardless of what the latent variable exactly represents. That is, we are interested in recovering the correct causal effect and not the true hidden confounder. Our goal is then to study when the resulting causal effect estimates are \emph{consistent}. Consistency means in general that the estimates of the parameters of interest approach their correct values as the amount of data increases, see, e.g., \citep{schervish2012theory}. 

For consistency to be possible, the model also has to be \emph{identifiable} with respect to causal effect estimates, so that minimizing the loss (e.g. maximizing the likelihood or the ELBO) does not map to multiple data generating parameters $\theta$ that correspond to different $p_\theta(y|do(t))$, see, e.g., \citep{murphy2012machine}. This is not to be confused with the common usage of the word "identifiability" in causal inference literature, where the question is instead whether it is possible in principle to estimate some causal effect from an observed distribution, e.g., using the techniques of the famous \emph{do-calculus} \citep{pearl2009causality}. In the presence of unobserved variables, results on causal identifiability often rely on parametric knowledge about the underlying data generating process. We refer to the identifiability of CEVAE as a statistical estimator as the \emph{model identifiability}, to distinguish it from the causal identifiability. Note that model identifiability as defined here guarantees only that a unique causal effect estimate is obtained, but not necessarily that it is consistent, for example if the model is misspecified.



The original CEVAE paper suggested multiple scenarios as the motivation for the model, including the case where we have very few parametric assumptions about the data generating process and the case where the distribution of the proxies is complex, as is often reasonable to assume with real-world data. Our aim here is to study how well the model works as we reduce the number of assumptions and move further into the territory envisioned in the paper. Based on the motivating scenarios, we can conceptually separate three goals for CEVAE:
\begin{enumerate}
    \item It should produce correct estimates with minimal knowledge about the parametric forms of the data generating $p(x|z)$, $p(t|z)$ and $p(y|z,t)$ distributions.
    \item It should work if we don't know the form of the unobserved confounder's distribution, which could be categorical or Gaussian, for instance.
    \item It should work with an arbitrarily complex distribution of proxies.
\end{enumerate}
With the first goal, the hope is that neural networks will estimate the conditionals correctly enough to estimate the causal effects. With the second goal, the hope is that the true confounder is represented well enough with the standard Gaussian prior of the VAE.


In practice, correct estimation could be prevented by many factors. For real-world data sets, it is possible that the causal effect is not identifiable at all from the data, even if we know the parametric form of the data generating process. If it is identifiable in principle, CEVAE might still fail due to inherent model nonidentifiability caused by the nonparametric assumptions and because we don't have a guarantee that finding a unique global optimum for the ELBO leads to a correct causal estimate. Local minima or other issues with optimization could cause further practical problems with the correctness of the results. 

We focus on the estimation of $p(y|do(t))$ instead of individual-level causal effects $p(y|do(t),x)$ as that makes the analysis more straightforward. Note that while the average treatment effect (ATE), defined as $\mathbb{E}[y|do(t=1)]-\mathbb{E}[y|do(t=0)]$, is a common metric, we are interested in $p(y|do(t))$ directly because the estimated ATE can have the correct value even when the estimates of $p(y|do(t=1))$ and $p(y|do(t=0))$ are not correct, and furthermore our analysis extends to continuous $t$.

\section{Results}

\subsection{Setup with provably identifiable simple synthetic data}

As a first step, we studied the simplest possible cases where we know from theory that the causal effects are identifiable in principle and provably correct analytical estimation methods exist, i.e., the two data types discussed in \citep{kuroki2014measurement}. Aside from being important basic cases, these data sets are interesting because they are relatively simple to study, we can compare CEVAE results to the analytical methods, and we can try to extract some qualitative understanding from the results. 

\textbf{Linear-Gaussian data} The first data type was the linear-Gaussian, where the true data generating distribution is such that all variables are jointly normally distributed, but respecting the conditional independences of graph 2 in Fig.\ref{fig:causal_graphs}b. The proxy $x$ also consists of two variables $x_1$ and $x_2$ that are conditionally independent given z:
\begin{gather*}
    z \sim N(0,1), x_1|z \sim N(c_1 z, \sigma_{x_1}), x_2|z \sim N(c_2 z, \sigma_{x_2})\\
    t|z \sim N(c_t t, \sigma_t), y|z,t \sim N(c_{yz} z + c_{yt} t, \sigma_y)
\end{gather*}
Here, $c_i$ and $\sigma_i$ are predefined parameters. To avoid jumping to conclusions based on arbitrary choices of data generating parameters, we sampled them randomly from a distribution that should provide a wide range of different generating processes. We detail the sampling method in the Supplementary Material. 


\textbf{Binary data} In the other type of data, all variables were binary, and in particular, $x$ consisted of two binary variables $x_1$ and $x_2$ that were conditionally independent given $z$. We sampled the data generating process from a distribution explained in the Supplementary Material. The binary data also tests the ability of CEVAE to perform correctly even if the assumption of a normally distributed unobserved confounder is not valid. 

To estimate the correctness of causal effect estimates using neural networks with the linear-Gaussian data, standard metrics such as ATE error do not apply because the treatment variable is not binary. Instead, we define the \emph{Average Interventional Distance} (AID):
\begin{equation}
    \textrm{AID} = \int p(t) \int |p_\theta(y|do(t)) - p(y|do(t))| dy dt
\end{equation}
where the integrals can be changed to sums for discrete variables. In addition to being defined for continuous treatments, this metric has the advantage that it will only approach zero if $p_\theta(y|do(t))$ approaches $p(y|do(t))$ for all values of $t$ that we have data from, and it follows the intuition that we should be more confident for values of $t$ for which we have lots of data.


\textbf{Estimation models} In the experiments, we refer to the "full CEVAE" as a model where all conditional distributions are parameterized with three-layer MLPs with layer width 30 and a latent variable dimension of 10. With the linear-Gaussian data, the standard deviations of the Gaussian conditionals were individually estimated for each data point, using the standard assumption of diagonal covariance in the encoder and decoder. For the linear-Gaussian data, we also considered the "linear CEVAE", a model with conditional distributions represented by simple linear layers and single standard deviations, shared between all data points and estimated for each conditional. We tried latent dimensionalities of 1, 2, and 10 for these linear models. Further details are provided in the Supplementary Material.


\subsubsection{Results with linear-Gaussian data}

It is, in fact, possible to show that the one-dimensional linear CEVAE is consistent in this situation, as encapsulated in the following proposition:
\begin{prop}
    A linear CEVAE with a one-dimensional latent space estimates the causal effect correctly, given that it reaches the global optimum of the ELBO with infinite data.
\end{prop}
The proof, which relies on the earlier result by \citep{kuroki2014measurement}, is provided in the Supplementary Material. Thus, a well-specified CEVAE model can result in correct estimation. The question then remains whether overparameterization by neural networks breaks the consistency.

\begin{figure}[]
    \centering
    \includegraphics[width=\textwidth]{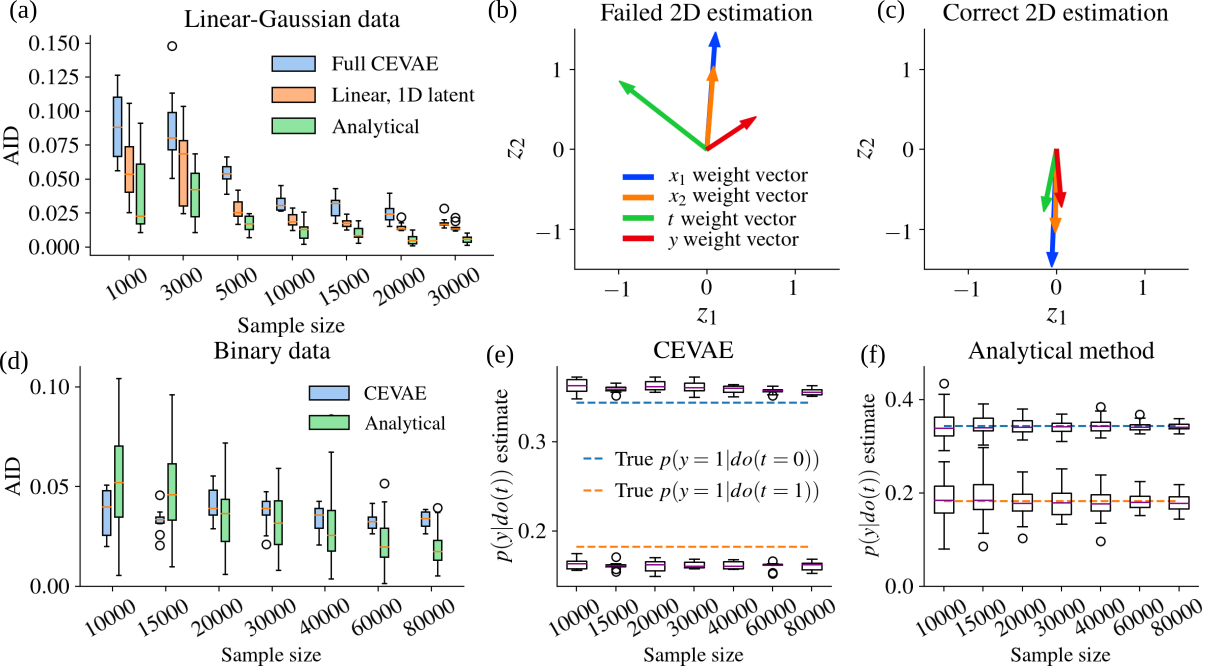}
    \caption{Top row, linear-Gaussian generative model: (a) AID values for the full CEVAE (NN conditionals and 10D latent space), for the simpler CEVAE with linear conditionals and 1D latent space, and for the analytical method. (b) Weight vectors of the different conditional distribution mean functions for a linear CEVAE with a two-dimensional latent space. Here parameters were initialized manually, and estimation failed. (c) The same parameter estimates for a model that estimated causal effects correctly. Here, only one latent dimension is used. Bottom row, binary generative model: (d) AID values for the full CEVAE and the analytical method with respect to sample size. (e) Estimates of the $p(y|do(t))$ values for full CEVAE. (f) The same estimates with the analytical method.}
    \label{fig:toydata}
\end{figure}

Figure \ref{fig:toydata}a shows the AID of the different models and the analytical method of \citep{kuroki2014measurement} as a function of sample size, for one data generating distribution. Analytical estimates for some of the required parameters for calculating $p(y|do(t))$ were not provided in the original paper, but we derive them in the Supplementary Material. While the analytical method and the linear CEVAE with a 1D latent variable perform better than the full CEVAE, all of them seem to converge towards the correct $p(y|do(t))$ distribution. We show in the Supplementary Material that the result is robust, as the estimates converge to the correct distribution for other data-generating parameters. Thus, overparameterizing the conditional distributions with NNs or using a larger than required latent variable dimension doesn't necessarily break the estimation of the causal effect. 

To highlight that the result is not obvious, we ran additional experiments with a model using linear conditional distributions but with a two-dimensional latent space, i.e., with one redundant dimension. With some initializations, the model ended up estimating the causal effect incorrectly, but with an indistinguishable ELBO compared to a model with the correct causal effect. The initialization and other details are given in the Supplementary Material. Figure \ref{fig:toydata}b visualizes the minimum with the wrong result. Essentially, the model uses only one latent dimension to reconstruct the proxies $x$ while treatment $t$ and outcome $y$ are reconstructed partly with the other dimension as well, preventing correct modeling of dependencies between observed variables. In practice, with a random initialization, this happens only rarely and not at all with the full CEVAE due to the tendency of \textit{posterior collapse} in VAEs, causing the model to use one dimension only. An example for the 2D linear CEVAE is shown in Fig.\ref{fig:toydata}c. Hence, whereas the posterior collapse is often an unwanted characteristic of VAEs~\citep{he2018lagging, razavi2019preventing, dai2020usual}, it here seems to save the day, although an unnecessarily high latent dimension could still cause issues in principle. In the Supplementary Material, we visualize the posterior collapse phenomenon for the full CEVAE and also show that the 10D linear CEVAE causal effect estimates become systematically incorrect if we prevent posterior collapse using KL divergence annealing.



\textbf{Conclusion} Overparameterizing the conditionals with neural networks does not necessarily prevent correct estimation, but a too high-dimensional latent variable could in principle. Posterior collapse usually resolves the problem, however.

\subsubsection{Results with binary data}

The AID values and corresponding causal effect estimates for the full CEVAE and the analytical method by \citep{miao2018identifying} are plotted as a function of sample size in Fig.\ref{fig:toydata}d. CEVAE, which incorrectly assumes a Gaussian latent variable, produces reasonable results but fails to converge to the correct causal effect. In contrast, when using the analytical method, the estimate gets better and better as the sample size increases. We show in the Supplementary Material that similar results are obtained for different data generating distributions. Note that the binary data generating process is possibly the simplest process where the actual confounder is not normally distributed and the causal effects are identifiable in principle. We conclude that CEVAE does not, in general, estimate the causal effect correctly when the latent variable is not specified appropriately in advance, and the second goal of CEVAE, mentioned in Sec.\ref{sec:estimability}, is not met. In the Supplementary Material, we also show that a version of CEVAE with a binary latent variable can produce correct causal estimates, although it is prone to get stuck in local minima for some data sets.


\textbf{Conclusion} CEVAE does not estimate causal effects correctly if the latent variable is misspecified, in general. 

\subsection{Illustration of difficulties with complex data}
This section describes two additional experiments with synthetic data, illustrating issues that can reasonably be assumed to come up with real data as well. The causal effects are identifiable with these data sets because they are based on the linear-Gaussian data. The estimation model is the same as the full CEVAE as specified in the previous section unless mentioned otherwise.

\subsubsection{Data with irrelevant variation in the proxies}

\begin{figure}
    \centering
    \includegraphics[width=\linewidth]{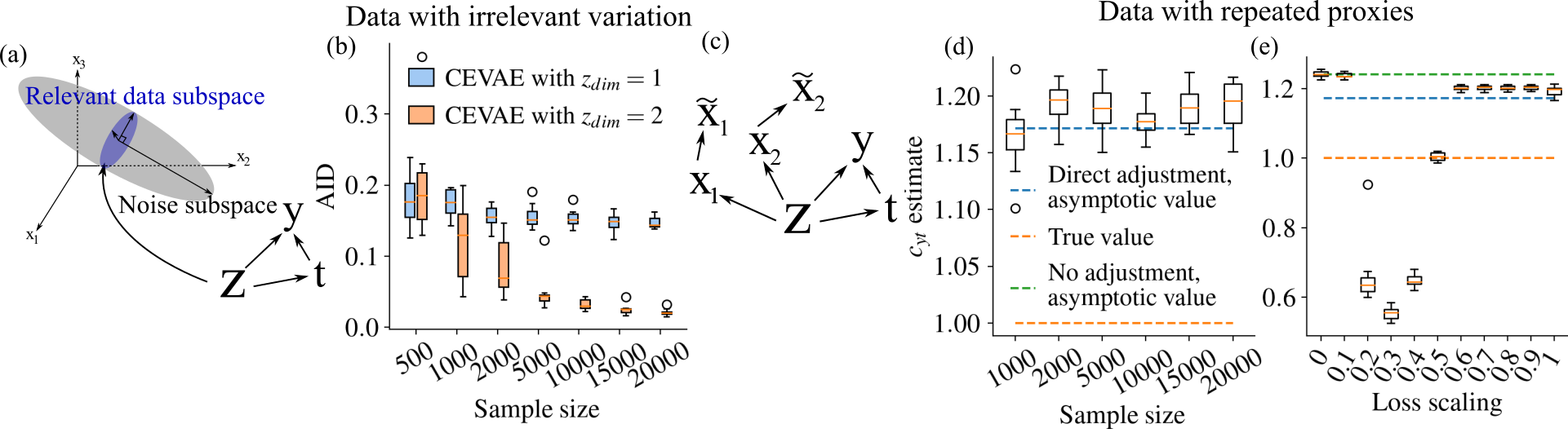}
    \caption{(a) Linear-Gaussian data generating process with irrelevant noise and a "hidden" relevant subspace. (b) AID values with respect to sample size for models using 1D and 2D latent spaces for the data with irrelevant noise. (c) Data generating process for linear-Gaussian data with copies $\tilde x_1$ and $\tilde x_2$. (d) Linear, 1D CEVAE $c_{yt}$ estimate as a function of sample size for the linear-Gaussian data with copies. (e) The $c_{yt}$ estimates as a function of proxy loss scaling for the linear 1D CEVAE.}
    \label{fig:complex_problems}
\end{figure}

The former experiments with linear-Gaussian data showed that an unnecessarily high-dimensional latent variable could cause issues, especially with the simpler linear CEVAE which did not exhibit posterior collapse, but the problem could be avoided by using a 1D latent. To see how a higher-dimensional latent variable could be necessary with complex data, we used the linear-Gaussian data set but added a third proxy variable that contained irrelevant, high-variance noise. After generating the proxies, an additional rotation was applied to them in the three-dimensional space so that the relevant variation was "hidden" in a specific two-dimensional subspace. The process is illustrated in Fig.\ref{fig:complex_problems}a. Here, with the standard assumption of diagonal covariance in the decoder, we can expect a 1D CEVAE to focus on modeling the noise because that dominates the loss. In contrast, a higher-dimensional CEVAE has the option to explain the noise with one and the signal with another dimension, enabling correct inference. 

Figure \ref{fig:complex_problems}b shows the full CEVAE AID as a function of sample size for one- and two-dimensional latent variables. As expected, the model with a two-dimensional latent variable estimates the causal effect correctly, while the one-dimensional model does not.

\textbf{Conclusion} A high-dimensional latent variable can be necessary to estimate the causal effect correctly using very noisy proxies, even if the actual confounder is one-dimensional.

\subsubsection{Data with repeated proxies}\label{sec:repeated_proxies}

The second difficulty in complex data can arise when the proxies contain variation relevant to predicting $t$ and $y$, but there exist significant correlations in $x$ that are not caused by the confounder. This can result in the model adjusting directly to $x$, ignoring the unobserved confounder entirely. As an example, consider the linear-Gaussian data set, but with a modification that we add two proxies, $\tilde x_1$ and $\tilde x_2$, which are copies of the original two. Here, the following proposition holds:
\begin{prop}
    With an altered linear-Gaussian data generating process where we have additional proxies $\tilde x_1 = x_1$ and $\tilde x_2 = x_2$, the value of the ELBO of a 2D CEVAE can approach infinity while the causal effect estimate converges to the value that is obtained by adjusting directly to the proxies, that is, $\int p(y|x,t)p(x)\textrm{d}x$.
\end{prop}
The proof is included in the Supplementary Material. The intuition is that whenever the model reconstructs one of the proxies, it can easily reconstruct the copy as well with the same accuracy, effectively doubling the importance of the proxy reconstruction loss. If we use the latent representation to directly represent the proxies with increasing accuracy, the negative KL divergence term in the ELBO decreases slower than the proxy reconstruction term increases, and the ELBO approaches infinity. At the same time, the latent space becomes a representation of the proxies, and the $y$ reconstruction term in the ELBO forces the corresponding predictor $p_\theta(y|z,t)$ to become an approximation of $p(y|x,t)$. We hypothesize that the same phenomenon will be an issue with a complicated distribution of proxies since, most likely, there will be similar correlations that are not directly caused by the unobserved confounder and that can cause the model to focus too much on proxy reconstruction. 

As a simple empirical demonstration of this phenomenon in a more realistic distribution, we experimented with data where a small Gaussian noise is added to $\tilde x_1$ and $\tilde x_2$ so that the correlations with $x_1$ and $x_2$ are not perfect. We modified the full CEVAE to use a linear predictor for the conditional distribution of $y$ to make the results easier to interpret (the estimated regression coefficient of $t$ should approach the $c_{yt}$ coefficient in the data generating process, i.e., the true causal effect). Figure \ref{fig:complex_problems}d shows the $c_{yt}$ estimate as a function of sample size. As expected, the estimate corresponds to the direct adjustment value that we would get if we used linear regression to predict $y$ based on $x$ and $t$.

Given that the loss function with repeated proxies corresponds to the regular loss where the proxy reconstruction term is multiplied by a factor of two, the most obvious way to resolve the problem is to adjust the reconstruction loss manually by a factor of one-half. In general, we would scale the term $\mathbb{E}_{q_\phi(z|x^i,t^i,y^i)}[\log p_\theta(x^i|z)]$ in Eq.\ref{eq:ELBO} by some factor $\lambda < 1$, forcing the VAE to put less weight on just reconstructing the proxies. However, in a real-world situation, it is not obvious what this scaling factor should be. To illustrate the effect of the scaling factor, Figure \ref{fig:complex_problems}e shows the $c_{yt}$ estimate as a function of proxy loss scaling for a sample size of 20000. With scaling factors close to one, the results are close to the direct adjustment results for the reasons explained. When lowering it to one-half, the estimate abruptly changes and ends up in the true value, as expected. When lowering it further, however, the results change as well and we start getting incorrect values. As the scaling factor approaches zero, CEVAE stops using the proxy data at all, and our estimate for $c_{yt}$ becomes equal to the one we would get by assuming $p(y|do(t))=p(y|t)$, i.e., no confounding. In the Supplementary Material, we prove the following proposition, which states that this estimate corresponds to many global optima of the ELBO where the latent space is either neglected entirely or not used in the reconstruction of either $t$ or $y$:
\begin{prop}
    With the proxy reconstruction loss scaled to zero, one set of global optima to the CEVAE ELBO is such that $p_\theta(y|do(t)) = p(y|t)$ and either $t$ or $y$ is not dependent at all on the latent variable for the linear-Gaussian data. 
\end{prop}
The intuition is that if one of the causal links $z\rightarrow t$ or $z\rightarrow y$ is removed in CEVAE, then the $c_{yt}$ estimate is produced as if there was no confounding. While other global optima exist, it seems that we get these no-adjustment solutions in practice. 

\textbf{Conclusion} CEVAE can overemphasize modeling of the proxies with some data sets, leading it to ignore the unobserved confounder entirely. We may be able to fix this by scaling the reconstruction loss for proxies, but it is not clear how to choose the scaling in practice. 

\subsection{Semi-synthetic data}

Here we describe two experiments based on real-world data sets. Details on the experimental setup, data generating processes, and neural network architectures are provided in the Supplementary Material. Here, we don't have guarantees that the causal effects are identifiable in principle from the data, which corresponds to the real-world situation where we can never be sure about identifiability without access to the parametric form of the data generating process. In any case, the experiments allow us to highlight issues with estimation failure that are not related to the identifiability of the data. 



\subsubsection{Proxy IHDP data set}


We decided to investigate the performance of CEVAE on a modified version of the Infant Health and Development Program (IHDP) data set~\citep{hill2011bayesian}, which was created from a study on the effects of intensive child care on future test scores of premature infants~\citep{brooks1992effects}. It consists of 25 covariates, some continuous and some categorical, treatments, and synthetically generated test scores. The IHDP data is a well-known causal inference benchmark, but as such, it is not suitable for our study because the $y$ values in the data have been generated directly based on the covariates, and thus the data doesn't follow the causal graph assumed by CEVAE. To overcome this, we singled out one of the covariates to be the hidden confounder, leaving the rest as proxies. The treatment $t$ and the recovery $y$ were then generated using the chosen confounder. Technically speaking, we don't know the direction of causalities between the confounder and the proxies defined this way, but the data is still Markov equivalent to the graph in Fig.\ref{fig:causal_graphs}b. Since the original data was very small (just 747 subjects), we trained a variational autoencoder on the covariates to generate more data from a distribution that is similar to the original one and which should be realistic enough for our purposes. The generated data set has the benefit that the distribution of the unobserved confounder and the proxies is not arbitrarily defined, instead following a real-world distribution that is relevant for causal inference. In the Supplementary Material, we show that the unobserved confounder is clearly correlated with many of the proxies, and thus it's possible that the proxies provide us enough information to make the causal effects identifiable in principle.


Figure \ref{fig:ihdp_mnist_results}a shows that the causal estimate does not approach the correct value as we increase the sample size, instead corresponding to the value we would get from direct adjustment to proxies, similarly to Section~\ref{sec:repeated_proxies}. Note that although we don't have strict guarantees that the causal effects are identifiable with this data set, the result nevertheless shows that the problem of placing too much weight on proxy reconstruction is relevant in a realistic use-case as well. In Fig.\ref{fig:ihdp_mnist_results}b we investigate whether scaling the proxy loss can recover the correct causal effect. The pattern is similar as before: Scaling with a factor close to one recovers direct adjustment. With a scaling factor close to zero, $y$ is predicted solely based on $t$. With intermediate scaling, some estimates are approximately correct, but this time not consistently for any of the scaling factors.


\textbf{Conclusion} The problem of adjusting directly to the proxies, described in Sec.\ref{sec:repeated_proxies}, happens with real data.

\subsubsection{Proxy MNIST data set}


\begin{figure*}[t]
    \centering
    \includegraphics[width=\linewidth]{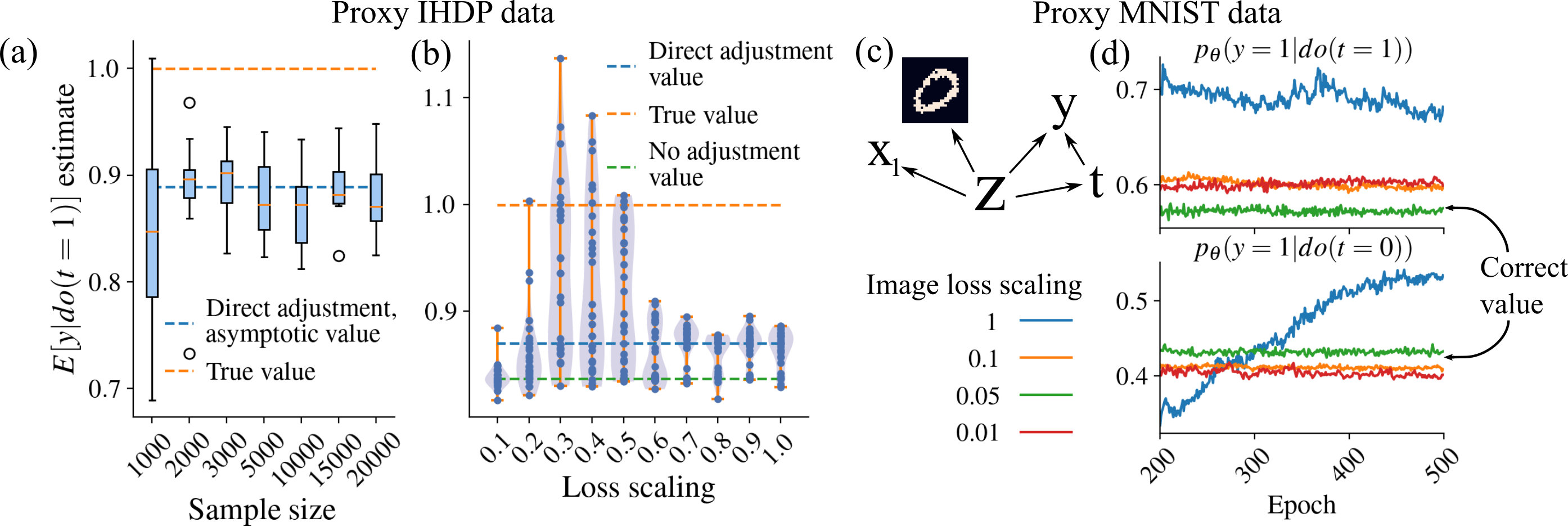}
    \caption{(a) $\mathbb{E}[y|do(t=1)]$ estimates for the IHDP data with respect to sample size. (b) $\mathbb{E}[y|do(t=1)]$ estimates for the IHDP data as a function of proxy loss scaling, for a data set of size 20000. The figures with $\mathbb{E}[y|do(t=0)]$ are included in the Supplementary Material. (c) The data generating process for the proxy MNIST data set. (d) Causal effect estimates for different image loss scaling values with respect to training time, for a data set of size 10000.}
    
    
    \label{fig:ihdp_mnist_results}
\end{figure*}

In this section, we experiment with a data set where the proxies are images, which is a natural domain for neural network based models such as CEVAE. To create data that follows the correct causal graph, we trained a GAN with a three-dimensional latent variable on the MNIST data~\citep{lecun-mnisthandwrittendigit-2010}, after which we used the GAN to generate a value between zero and one for each pixel given the latent value. We interpreted these as probabilities and sampled each pixel from the corresponding Bernoulli distribution to generate noisy images, which were used as the proxies. We used the first latent dimension of the GAN as the unobserved confounder $z$ and generated binary $t$ and $y$ values based on the chosen $z$ for each sampled image. We also generated an additional linear-Gaussian proxy $x_1$ to increase the chance that the causal effects are identifiable in principle, while not reducing the problem to trivial. The created data set, illustrated in Fig.\ref{fig:ihdp_mnist_results}c, then followed the correct causal graph with a normally distributed unobserved confounder. 



Since the image data is very complex and high-dimensional, it is reasonable to expect that CEVAE can put too much weight on image reconstruction in one way or another during training. Thus, we experimented by scaling the reconstruction loss term of the images to different values. Indeed, figure \ref{fig:ihdp_mnist_results}d shows us that without any scaling, the estimated $p(y|do(t))$ values do not converge to anything even with 500 epochs. This is possibly due to the reconstruction term being much larger than the rest, so that fluctuations there overshadow the modeling of the other variables. When scaling the image loss to lower values, the estimates start to converge but are not quite correct for values 0.1 and 0.01. For scaling value 0.05, however, we recover almost exactly the correct result. The result is confirmed in the Supplementary Material with AID values from multiple runs. Thus, it seems that scaling the loss function appropriately can result in the correct causal effect even with data as complex as images, if we know the correct scaling.


\textbf{Conclusion} With very complex proxy data, getting an estimate for the causal effect can be difficult. In some cases, scaling the loss appropriately can result in correct estimation even with real data. 

\subsubsection{Twins data set}

We also ran experiments on the Twins data set presented in the original paper~\citep{louizos2017causal, nber_twins}. It provides an example where the causal effects are identifiable in principle, the confounder is categorical, and where the distribution of $z$,$t$, and $y$ is based on real-world data. In the Supplementary Material, we show that CEVAE doesn't return consistent estimates, further reinforcing our conclusion about incorrect estimates with misspecified latent variables.

\section{Discussion}\label{sec:discussion}


Two of the goals we listed for CEVAE in Section~\ref{sec:estimability} were that it should recover causal effects even if we don't know the distribution of the confounder and if the distribution of the proxies is complex. It appears that CEVAE does not work consistently correctly in either case. Thus, while using a deep latent variable model in this context shows some promise, new solutions are needed to overcome the issues that come up with real data.  Although there is an absence of theory supporting the model in general, these results were non-obvious to us and we believe that they are useful for many others, given the large amount of research in the field. The main limitation of our work is that we focused on CEVAE, but we believe that the qualitative results and recognized problems will be useful in research on other, similar models. Another limitation is that our study was mainly empirical and we can not provide theoretical guarantees that the qualitative results transfer to all possible data sets. The negative results do, however, serve as counterexamples, and we did attempt to provide intuition for the phenomena observed, allowing future researchers to assess whether our work is relevant for them.

Finally, our aim is not to discourage research with CEVAE or deep latent variable models for causal inference in general, but instead, we hope that our results will accelerate progress in the field. The hope is that our results are a starting point for thinking about the consistency of causal estimation with similar models and advancing guarantees for it. In any case, our experiments showed that we should not brush off the issue entirely since the alternative is that the model can produce incorrect results even with some rudimentary data sets. 

\section{Impact Statement}\label{sec:impact}

If CEVAE and similar machine learning based methods for causal inference become usable and prevalent in application areas such as epidemiology and the social sciences, this line of work could have a clear positive social impact by enabling research in these fields with new possibilities and thus allowing for better decision-making through better understanding of important phenomena. On the other hand, it's possible that practitioners become overly reliant on the claims that these methods estimate causal effects with very few assumptions, becoming less rigorous in considering the assumptions that are necessary to make (e.g., the causal graph). This could inadvertently have the negative impact of degrading the quality of research. Researchers in machine learning and causal inference should strive to avoid this by communicating the limitations realistically to practitioners and make the necessary assumptions as explicit as possible. 

\section*{Acknowledgements}

PM has received funding from the Academy of Finland (grants 336033, 315896), BusinessFinland (grant 884/31/2018), and EU H2020 (grant 101016775).



\bibliography{example_paper}
\bibliographystyle{abbrvnat}

\clearpage

\appendix

\maketitle

\section{Analytical estimate of $p(y|do(t))$ for linear-Gaussian data}\label{sec:pydot_estimate}

In this section, we derive an analytical estimate of the parameters of the $p(y|do(t))$ distribution for the linear-Gaussian data, including the ones that weren't provided in the original paper. It assumes that we know the parametric form of the generating process. The approach is slightly different from the original paper, which utilized higher-level properties of the structural model in their derivation. We will first derive the asymptotic, infinite data covariance matrix of the observed variables expressed using the data generating parameters $c_1,c_2,\sigma_1$, etc., after which we can derive expressions for the parameters using observable covariances. The formulas can then be used as asymptotically correct estimates with finite data as well.

We start by finding out a form for the joint distribution, including $z$:
\begin{align}
    p(z,x,t,y) &= p(z)p(x|z)p(t|z)p(y|z,t)\nonumber\\
    &\sim e^{-\frac{1}{2} \left(z^2 + \frac{(c_1 z - x_1)^2}{\sigma_1^2} + \frac{(c_2 z - x_2)^2}{\sigma_2^2} + \frac{(c_t z - x_t)^2}{\sigma_t^2} + \frac{(c_{yz}z + c_{yt}t - y)^2}{\sigma_y^2}\right)}\nonumber
\end{align}
This is a jointly Gaussian distribution. We can find the covariance matrix by looking at the exponent and rearranging terms:
\begin{align*}
    -2\log p(z,x,t,y) &\sim z^2(1+ \frac{c_1^2}{\sigma_1^2} + \frac{c_2^2}{\sigma_2^2} + \frac{c_t^2}{\sigma_t^2} + \frac{c_{yz}^2}{\sigma_y^2}) + zx_1(-\frac{2c_1}{\sigma_1^2}) + \\ &zx_2(-\frac{2c_2}{\sigma_2^2}) + zt(-\frac{2c_t}{\sigma_t^2} + \frac{2c_{yz}c_{yt}}{\sigma_y^2}) + zy(-\frac{2c_{yz}}{\sigma_y^2}) + x_1^2\frac{1}{\sigma_1^2} + \\&x_2^2\frac{1}{\sigma_2^2} + t^2(\frac{1}{\sigma_t^2} + \frac{c_{yt}^2}{\sigma_y^2}) + ty(-\frac{2c_{yt}}{\sigma_y^2}) + y^2(\frac{1}{\sigma_y^2})\\
    &= \begin{bmatrix} z & x_1 & x_2 & t & y \end{bmatrix} C^{-1} \begin{bmatrix} z \\ x_1 \\ x_2 \\ t \\ y \end{bmatrix}
\end{align*}
where
\begin{align*}
    C^{-1} = \begin{bmatrix} 1+ \frac{c_1^2}{\sigma_1^2} + \frac{c_2^2}{\sigma_2^2} + \frac{c_t^2}{\sigma_t^2} + \frac{c_{yz}^2}{\sigma_y^2} & -\frac{c_1}{\sigma_1^2} & -\frac{c_2}{\sigma_2^2} & -\frac{c_t}{\sigma_t^2} + \frac{c_{yz}c_{yt}}{\sigma_y^2} & -\frac{c_{yz}}{\sigma_y^2} \\
    -\frac{c_1}{\sigma_1^2} & \frac{1}{\sigma_1^2} & 0 & 0 & 0 \\
    -\frac{c_2}{\sigma_2^2} & 0 & \frac{1}{\sigma_2^2} & 0 & 0 \\
    -\frac{c_t}{\sigma_t^2} + \frac{c_{yz}c_{yt}}{\sigma_y^2} & 0 & 0 & \frac{1}{\sigma_t^2} & -\frac{c_{yt}}{\sigma_y^2} \\
    -\frac{c_{yz}}{\sigma_y^2} & 0 & 0 & -\frac{c_{yt}}{\sigma_y^2} & \frac{1}{\sigma_y^2}
    \end{bmatrix}
\end{align*}
Inverting this, we get $C$, and the covariance matrix $C_{xty}$ of the marginal distribution $p(x,t,y)$ is got by dropping the row and columns corresponding to $z$, since $p(z,x,t,y)$ is jointly Gaussian:
\begin{equation}
\footnotesize
\setlength{\arraycolsep}{2.5pt}
\medmuskip = 1mu 
    C_{xty} = \left[
\begin{array}{cccc}
  {c_1}^2+{\sigma_1}^2 & {c_1} {c_2} & {c_1} {c_t} & {c_1} ({c_t} {c_{yt}}+{c_{yz}}) \\
 {c_1} {c_2} & {c_2}^2+{\sigma_2}^2 & {c_2} {c_t} & {c_2} ({c_t} {c_{yt}}+{c_{yz}}) \\
 {c_1} {c_t} & {c_2} {c_t} & {c_t}^2+{\sigma_t}^2 & {c_t}^2 {c_{yt}}+{c_t} {c_{yz}}+{c_{yt}} {\sigma_t}^2 \\
 {c_1} ({c_t} {c_{yt}}+{c_{yz}}) & {c_2} ({c_t} {c_{yt}}+{c_{yz}}) & {c_t}^2 {c_{yt}}+{c_t} {c_{yz}}+{c_{yt}} {\sigma_t}^2 & {c_t}^2 {c_{yt}}^2+2 {c_t} {c_{yt}} {c_{yz}}+{c_{yt}}^2 {\sigma_t}^2+{c_{yz}}^2+{\sigma_y}^2 \\
\end{array}
\right] \nonumber
\end{equation}
We then have a system of 10 equations, where each of the matrix cells corresponds to an asymptotic, infinite-data covariance. The equations of interest to us are
\begin{align}
    c_1 c_2 =\textrm{Cov}(x_1, x_2),\quad c_1 c_t = \textrm{Cov}(x_1, t)&,\quad c_2 c_t = \textrm{Cov}(x_2, t), \quad c_t^2 + \sigma_t^2 = \textrm{Var}(t) \nonumber \\
     c_2(c_t c_{yt} + c_{yz}) = \textrm{Cov}(x_2, y)&,\quad{c_t}^2 {c_{yt}}+{c_t} {c_{yz}}+{c_{yt}} {\sigma_t}^2 = \textrm{Cov}(t,y)\nonumber\\
     {c_t}^2 {c_{yt}}^2+2 {c_t} {c_{yt}} {c_{yz}}+{c_{yt}}^2 &{\sigma_t}^2+{c_{yz}}^2+{\sigma_y}^2 = \textrm{Var}(y)\nonumber
\end{align}
These can be solved to get
\begin{align}
    c_{yt} &= \frac{\textrm{Cov}(t,y) \textrm{Cov}(x_1,x_2) - \textrm{Cov}(x_2,y) \textrm{Cov}(x_1,t)}{\textrm{Var}(t) \textrm{Cov}(x_1,x_2) - \textrm{Cov}(x_1, t) \textrm{Cov}(x_2, t)}\\
    c_{yz}^2 &= \frac{\textrm{Cov}(x_1,t)\textrm{Cov}(x_1,x_2)(\textrm{Cov}(t,y)\textrm{Cov}(x_2,t) - \textrm{Var}(t)\textrm{Cov}(x_2,y))^2}{\textrm{Cov}(x_2,t)(\textrm{Var}(t)\textrm{Cov}(x_1,x_2) - \textrm{Cov}(x_1,t)\textrm{Cov}(x_2,t))^2}\\
    c_t^2 &= \frac{Var(t)\textrm{Cov}(x_2,t)}{\textrm{Cov}(x_1,x_2)}\\
    \sigma_t^2 &= \textrm{Var}(t) - c_t^2\\
    c_tc_{yz} &= \textrm{Cov}(t,y) - c_{yt}\sigma_t^2 - c_t^2c_{yt}\\
    \sigma_y^2 &= \textrm{Var}(y) - c_{yz}^2 - c_{yt}^2 \sigma_t^2 - 2c_{yt}c_tc_{yz} - c_t^2c_{yt}^2
\end{align}
where earlier expressions can be plugged in to later ones (especially $\sigma_y^2$ doesn't simplify much). The quantities $c_{yt}$, $c_{yz}^2$ and $\sigma_y^2$ are enough to characterize $p(y|do(t))$, since
\begin{align}
    p(y|do(t)) &= \int_{-\infty}^\infty \frac{1}{\sqrt{2\pi}}e^{-\frac{z^2}{2}}  \frac{1}{\sqrt{2\pi\sigma_y}}e^{-\frac{(y-c_{yz}z-c_{yt}t)^2}{2\sigma_y}}\textrm{d}z\nonumber\\
    &=  \frac{1}{\sqrt{2\pi\sigma_{y|do(t)}}}e^{-\frac{(y-\mu_{y|do(t)})^2}{2\sigma_{y|do(t)}}}
\end{align}
where
\begin{align}
\mu_{y|do(t)} &= c_{yt} t \\
    \sigma_{y|do(t)} &= \sqrt{\sigma_y^2 + c_{yz}^2}
\end{align}
In practice, we can use the asymptotically correct equations as formulas for estimation with finite data. The difference is that we use sample covariances and variances, and the parameter estimates are naturally correct only with infinite data. 

\section{Proof of Proposition 1: The 1D linear CEVAE is consistent with linear-Gaussian data}

In section \ref{sec:pydot_estimate} we showed that the causal effect $p(y|do(t))$ is identifiable from linear-Gaussian data, and presented an asymptotically correct, analytical method for estimation. Here we consider the 1D linear CEVAE, which estimates the conditional distributions linearly and has a latent dimension of one, thus being parameterized in the same way as the data generating process. We show that it is guaranteed to estimate the correct causal effect as well, assuming that we find the global optimum of the ELBO with infinite data. The proof relies on three facts:
\begin{enumerate}
    \item As shown in Sec.\ref{sec:pydot_estimate}, the parameters of the data generating process required for identifying the causal effect match one-to-one with observed covariances.
    \item The CEVAE estimation model is defined so that the parameters match exactly with the data generating parameters and the prior is correctly specified as well.
    \item The variational approximation to the posterior distribution can correctly represent the true posterior in this case.
\end{enumerate}

\textbf{Proof.} We denote $p_\theta(\cdot)$ as the distribution induced by the VAE, e.g. $p_\theta(z,x,t,y) = p_\theta(x_1|z)p_\theta(x_2|z)p_\theta(t|z)p_\theta(y|z,t)p(z)$, where $p(z)$ is the zero mean, unit variance prior of the VAE. 

Since the joint distribution $p_\theta(z,x,t,y)$ induced by the model is jointly Gaussian with mean zero, we know from properties of multivariate normal distributions that $p_\theta(z|x,t,y)$ is Gaussian as well with a mean that is a linear function of $x_1$, $x_2$, $t$ and $y$ with zero bias and constant variance. Thus, as we set our variational approximation $q_\phi(z|x,t,y)$ to be similarly a Gaussian with mean being a linear function of the observed variables and estimate the variance as one shared parameter, it can represent the true posterior $p_\theta(z|x,t,y)$ with the right choice of parameters $\phi$. Thus, the global optimum of the ELBO also equals the global optimum of the marginal log-likelihood. In the limit of infinite data, maximizing the sum of marginal log-likelihoods becomes equivalent to maximizing
\begin{align}
    \int p(x,t,y) \log p_\theta(x,t,y)\textrm{d}z = \int p(x,t,y) \log \left(\frac{p_\theta(x,t,y)}{p(x,t,y)}p(x,t,y)\right)\textrm{d}x\textrm{d}t\textrm{d}y \nonumber \\
    = -KL[p(x,t,y)||p_\theta(x,t,y)] + \int p(x,t,y)\log p(x,t,y) \textrm{d}x\textrm{d}t\textrm{d}y \label{eq:asym_marg_likelihood}
\end{align}
where $p(x,t,y)$ is the true distribution of the data. Thus, since the parameter space of the linear VAE includes the true distribution, at the globally optimal ($\theta$,$\phi$) combination the KL divergence goes to zero and $p_\theta(x,t,y) = p(x,t,y)$, i.e., the marginal distribution of our model is the true distribution. Because the VAE parameterization was defined in the exact same way as the generative model, we can then go through the exact same steps as we did in Sec.\ref{sec:pydot_estimate}, and notice that the estimate of $p(y|do(t))$ has to be the one we get from the true distribution. Thus, the model estimates $p(y|do(t))$ correctly. $\square$

\section{Proof of Proposition 2: We can get an infinite ELBO with copied proxies}
In this section, we prove that we can get an infinite ELBO by using the latent space solely to reconstruct the proxies for linear-Gaussian data where the proxies are copied at least once. The proof assumes a linear CEVAE estimation model with a latent dimension of at least two, but the result is valid for a neural network parameterized CEVAE as well assuming that it can represent the same conditional distributions as the linear CEVAE. Given the universal approximation capabilities of neural networks, this is not a very radical assumption to make. The central idea in the proof is that we find a certain path in the parameter space which we then show to lead to an infinite evidence lower bound. It is constructed by mapping each value of $x$ to a corresponding position in the latent space, after which we let the encoder and decoder variances go to zero, forcing the reconstruction to become perfect. 

\textbf{Proof.} Recall that the ELBO for CEVAE can be written in the form
\begin{align}
\mathcal{L}(\theta,\phi) &= \sum_i\big[\mathbb{E}_{q_\phi(z|x^{i},t^{i},y^{i})}[\log p_\theta(x^{i}|z) + \log p_\theta(t^{i}|z) + \log p_\theta(y^{i}|z,t^{i})] -\nonumber\\ &KL[q_\phi(z|x^{i},t^{i},y^{i})||p(z)] \big]
\end{align}
Let us now consider the scenario where we are trying to estimate the causal effect from a data set containing $N$ copies of the original two proxies. We restrict the analysis to the part of the parameter space where the variational approximation depends only on the proxies and both proxies are reconstructed using only one of the dimensions:
\begin{align}
    q_\phi(z|x,t,y) &= \mathcal{N}(z|\mu_{z|x}, S_{z|x})\\
    \mu_{z|x} &= \begin{bmatrix}\gamma_{z_1} x_1\\ \gamma_{z_2} x_2 \end{bmatrix}\\
    S_{z|x} &= \begin{bmatrix} s_{z_1|x}^2 & 0 \\ 0 & s_{z_2|x}^2 \end{bmatrix}
\end{align}
The proxy distribution in the decoder is set to $p_\theta(\{x_i\}|z) = \mathcal{N}(x_i|\gamma_i z, s_i^2)^N$, where $\gamma_i$ and $s_i$ are shared parameters for $x_i$ and all its copies, denoted by the set $\{x_i\}$ and $N$ is the number of copies $|\{x_i\}|$. Note that we use $\gamma$ and $s$ to highlight that these are parameters of CEVAE, not the data generating distribution, where we used $c$ and $\sigma$. Let's focus on the proxy reconstruction term and the KL divergence terms for the first latent dimension - proxy copy group pair, $z_1$ and $\{x_1\}$. The reconstruction term for a single observation is
\begin{align}
    \mathbb{E}&_{q_\phi(z_1|x,t,y)}[\log p_\theta(\{x_1\}|z)]\nonumber\\
    &=\int \frac{1}{\sqrt{2\pi s_{z_1|x}}} \exp(-\frac{(z_1-\gamma_{z_1} x_1)^2}{2 s_{z_1|x}^2})\log \left[\left(\frac{1}{\sqrt{2\pi s_1}}\exp( \frac{(x_1-\gamma_1 z)^2}{2s_1^2})\right)^N\right]dz\\
    &=\int \frac{1}{\sqrt{2\pi s_{z_1|x}}} \exp(-\frac{(z_1-\gamma_{z_1} x_1)^2}{2 s_{z_1|x}^2}) N \left(\log \frac{1}{\sqrt{2\pi s_1}} - \frac{(x_1-\gamma_1 z)^2}{2s_1^2}\right)dz\\
    &= N \log \frac{1}{\sqrt{2\pi s_1}} - N \frac{\sqrt{s_{z_1}}}{2 s_1^2}\left( \gamma_1^2 s_{z_1}^2 + x_1^2(1-\gamma_1 \gamma_{z_1}) \right)
\end{align}
where the exponentiation by $N$ is due to the $N$ identical copies that are reconstructed using the same parameters. Due to the diagonal assumptions in the prior and variational approximation, the KL divergence breaks into two parts:
\begin{align}
    -KL&[q_\phi(z|\{x_1\},\{x_2\},t,y)||p(z_1)] \nonumber\\
    &= -\int \int \mathcal{N}(z_1|\gamma_{z_1} x_1, s_{z_1|x}^2) \mathcal{N}(z_2|\gamma_{z_2} x_2, s_{z_2|x}^2) \log\left(\frac{p(z_1)p(z_2)}{\mathcal{N}(z_1|\gamma_{z_1} x_1, s_{z_1|x}^2) \mathcal{N}(z_2|\gamma_{z_2} x_2, s_{z_2|x}^2)}\right)dz_1 dz_2\\
    &=-KL[\mathcal{N}(z_1|\gamma_{z_1} x_1, s_{z_1|x}^2)||p(z_1)] - KL[\mathcal{N}(z_2|\gamma_{z_2} x_2, s_{z_2|x}^2)||p(z_2)]
\end{align}
where the term relevant for parameters regarding $z_1$ and $x_1$ is
\begin{align}
    -KL[\mathcal{N}(z_1|\gamma_{z_1} x_1, s_{z_1|x}^2)||p(z_1)] = -\log\frac{1}{s_{z_1|x}} - \frac{s_{z_1|x}^2 + \gamma_{z_1}x_1^2}{2} + \frac{1}{2}
\end{align}
Bringing the two terms together and restricting the parameter space further so that $\gamma_1 \gamma_{z_1} = 1$ and $s_1^\frac{4}{5} = s_{z_1|x} = s$, we have
\begin{align}
    \mathbb{E}_{q_\phi(z_1|x,t,y)}&[\log p_\theta(x_1|z)] - KL[\mathcal{N}(z_1|\gamma_{z_1} x_1, s_{z_1|x}^2)||p(z_1)] \nonumber\\
    &= N \log \frac{1}{\sqrt{2\pi s_1}} - N \frac{\sqrt{s_{z_1}}}{2 s_1^2}\left( \gamma_1^2 s_{z_1}^2 + x_1^2(1-\gamma_1 \gamma_{z_1}) \right) -\log\frac{1}{s_{z_1|x}} - \frac{s_{z_1|x}^2 + \gamma_{z_1}x_1^2}{2} + \frac{1}{2}\\
    &= -\frac{5N}{8}\log s - N\frac{\gamma_1^2}{2} + \log s - \frac{s^2}{2} + \textrm{constants}
\end{align}
Let's now consider the two scenarios where $N=1$ and $N=2$ we let $s \rightarrow 0$. 

\textbf{N=1} The sum of the relevant, non-constant terms in the limit approaches minus infinity:
\begin{equation}
    \lim_{s\rightarrow0}\left(-\frac{5}{8}\log s - \frac{\gamma_1^2 }{2} + \log s - \frac{s^2}{2}\right) = \lim_{s\rightarrow0}(\frac{3}{8}\log s) + 0 = -\infty
\end{equation}
Thus, with no copies this approach clearly doesn't maximize the ELBO.

\textbf{N=2} The sum approaches infinity in the limit:
\begin{equation}
    \lim_{s\rightarrow0}\left(-\frac{10}{8}\log s - \gamma_1^2 + \log s - \frac{s^2}{2}\right) = \lim_{s\rightarrow0}(-\frac{1}{4}\log s) + 0 = +\infty
\end{equation}
Although we only focused on a single observation in the ELBO, the final expression is not actually dependent on the values of $x_1$, so the ELBO will go to infinity for all observations with this parameterization. We can do the exact same thing for the second group of proxies $\{x_2\}$ and the second latent variable $z_2$. We conclude that while exactly this approach might not be the fastest way to increase the ELBO during training, it is clearly possible to get an infinite ELBO by using the latent space solely to reconstruct the proxies as accurately as possible. $\square$

\begin{figure}
    \centering
    \includegraphics[width=\textwidth]{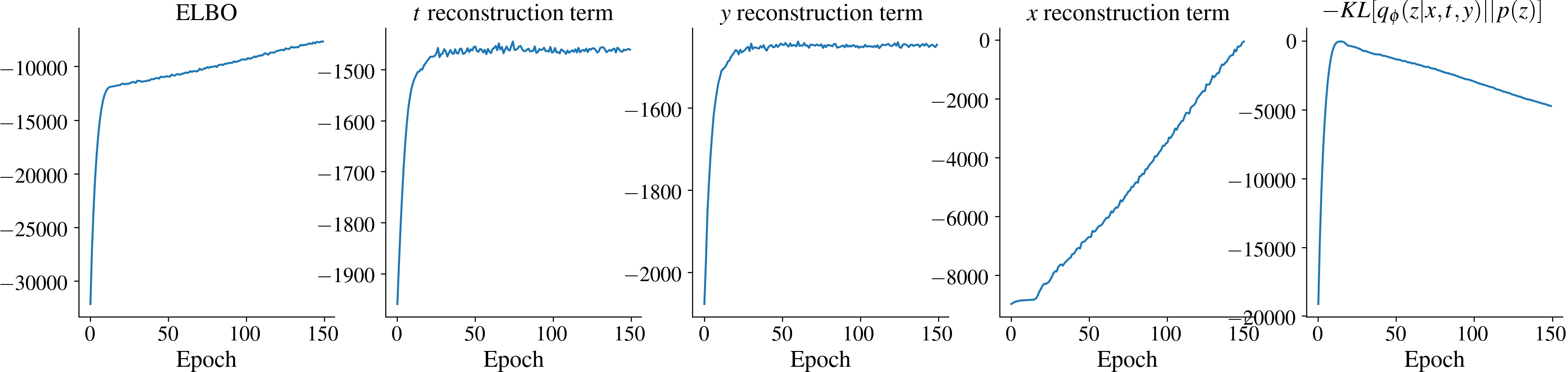}
    \caption{The ELBO and its parts for the linear-Gaussian data with exact proxies. The estimation model was the full, NN parameterized CEVAE. The data generating process was the same as in the main experiment, but without additional noise on the copies. The sample size was set to 1000, batch size was 200 and learning rate was 0.0001.}
    \label{fig:proxycopy_ELBO}
\end{figure}

In practice, the model could then follow a similar path as a result of gradient descent during training. Figure \ref{fig:proxycopy_ELBO} shows the ELBO and its parts for a full, neural network parameterized CEVAE with copied proxies. While $t$ and $y$ reconstruction terms in the ELBO converge very early, the $x$ reconstruction term keeps improving long after that. The negative KL divergence term gets smaller, but it is not enough to counter the increase in $x$ reconstruction quality. 

Intuitively, the reason that the final expression doesn't depend on the values of observed $x$ is that the latent space is able to represent the proxies perfectly, i.e., each $x_1$ is mapped to a corresponding latent representation through the $\gamma_{z_1}$ parameter. We restricted $\gamma_1\gamma_{z_1}=1$ because if the latent representation of the proxy was scaled down or up by $\gamma_{z_1}$, we need to do the opposite scaling $\gamma_1=\frac{1}{\gamma_{z_1}}$ in reconstruction. 

Note that if the latent dimension is larger than two, we won't improve the ELBO by using them to reconstruct the treatment $t$ and effect $y$, as shown in Sec.\ref{sec:analytical_posteriorcollapse}. Thus, the $p_\theta(t|z)$ and $p_\theta(y|z,t)$ reconstruction terms may only improve if the proxies, through the latent representation, are useful for predicting $y$ and $t$.

\section{Proof of Proposition 3: Posterior collapse in the 1D linear CEVAE with no proxies}\label{sec:analytical_posteriorcollapse}

Here we show analytically that if the proxy reconstruction term is set to zero (essentially, we don't have any proxies), then a set of solutions where $p_\theta(y|do(t))=p(y|t)$ are global maxima of the ELBO with linear-Gaussian data. These solutions correspond to situations where the latent space is not used in the reconstruction of $t$ or $y$, or either of them. In the proof, we assume that the estimation model is the CEVAE with a one-dimensional latent space and linearly parameterized conditionals. However, the result applies to a CEVAE with neural network parameterization as well if we assume that it can represent the same conditionals as the linear CEVAE. 

\textbf{Proof.} Let's assume that we have maximized the ELBO so that $q_\phi(z|x,t,y) = p_\theta(z|t,y)$ for whatever $\theta$ that can maximize it. Then, with infinite data, according to Eq.\ref{eq:asym_marg_likelihood}, we get that $p(t,y) = p_\theta(t,y)$. Let us use the notation $\gamma$ and $s$ to signify the CEVAE parameters that correspond to the parameters $c$ and $\sigma$ in the data generating model. In a similar way as in Sec.\ref{sec:pydot_estimate} we can then show that the inverse of the covariance matrix of $p_\theta(z,t,y)$ is then 
\begin{align*}
    C^{-1} = \begin{bmatrix} 1 + \frac{\gamma_t^2}{s_t^2} + \frac{\gamma_{yz}^2}{s_y^2} & -\frac{\gamma_t}{s_t^2} + \frac{\gamma_{yz}\gamma_{yt}}{s_y^2} & -\frac{\gamma_{yz}}{s_y^2} \\
    -\frac{\gamma_t}{s_t^2} + \frac{\gamma_{yz}\gamma_{yt}}{s_y^2} & \frac{1}{s_t^2} & -\frac{\gamma_{yt}}{s_y^2} \\
    -\frac{\gamma_{yz}}{s_y^2} & -\frac{\gamma_{yt}}{s_y^2} & \frac{1}{s_y^2}
    \end{bmatrix}
\end{align*}
Inverting and marginalizing w.r.t. $z$, we then get
\begin{equation}
    C_{ty} = \begin{bmatrix} {\gamma_t}^2+{s_t}^2 & {\gamma_t}^2 {\gamma_{yt}}+{\gamma_t} {\gamma_{yz}}+{\gamma_{yt}} {s_t}^2 \\
    {\gamma_t}^2 {\gamma_{yt}}+{\gamma_t} {\gamma_{yz}}+{\gamma_{yt}} {s_t}^2 & {\gamma_t}^2 {\gamma_{yt}}^2+2 {\gamma_t} {\gamma_{yt}} {\gamma_{yz}}+{\gamma_{yt}}^2 {s_t}^2+{\gamma_{yz}}^2+{s_y}^2 \end{bmatrix}\nonumber
\end{equation}
We have the equations 
\begin{align*}
    \textrm{Var}(t)&=\gamma_t^2 + s_t^2\\
    \textrm{Var}(y)&={\gamma_t}^2 {\gamma_{yt}}^2+2 {\gamma_t} {\gamma_{yt}} {\gamma_{yz}}+{\gamma_{yt}}^2 {s_t}^2+{\gamma_{yz}}^2+{s_y}^2\\
    \textrm{Cov}(t,y)&={\gamma_t}^2 {\gamma_{yt}}+{\gamma_t} {\gamma_{yz}}+{\gamma_{yt}} {s_t}^2
\end{align*}
This group of equations has many solutions, but two obvious groups of solutions stand out:

\textbf{Group 1. } $\gamma_{yz}=0, s_y^2=\textrm{Var}(y) - \textrm{Cov}(t,y)$ and $\gamma_{yt} = \frac{\textrm{Cov}(t,y)}{\textrm{Var}(t)}$.

This corresponds to the solution where $z$ doesn't have a direct causal effect on $y$, and thus there is no confounding and CEVAE doesn't use $z$ in reconstruction of $y$. We can also show that $p_\theta(y|do(t))=p(y|t)$, by first calculating $p_\theta(y|do(t))$:
\begin{align}
    p_\theta(y|do(t)) &= \int p_\theta(y|z,t)p(z)dz\nonumber\\
    &= p_\theta(y|z,t)=p_\theta(y|t)\nonumber\\
    &= \mathcal{N}(y|\gamma_{yt}t,s_y^2) = \mathcal{N}(y|\frac{\textrm{Cov}(t,y)}{\textrm{Var}(t)}t,\textrm{Var}(y) - \textrm{Cov}(t,y))
\end{align}
Here the second equality is true because $y$ is not dependent on $z$. The result corresponds to the conditional distribution formula for bivariate Gaussians: $p(y|t) = \mathcal{N}(y|\frac{\textrm{Cov}(t,y)}{\textrm{Var}(t)}t,\textrm{Var}(y) - \textrm{Cov}(t,y))$. 

\textbf{Group 2. } $\gamma_t=0, s_t^2=\textrm{Var}(t), \gamma_{yz}^2+s_y^2=\textrm{Var}(y) - \textrm{Cov}(t,y)$ and $\gamma_{yt} = \frac{\textrm{Cov}(t,y)}{\textrm{Var}(t)}$.

This corresponds to the solution where $z$ doesn't have a direct causal effect on $t$, and thus again there is no confounding. Again, we can calculate $p_\theta(y|do(t))$:
\begin{align}
    p_\theta(y|do(t)) &= \int p_\theta(y|z,t)p(z)dz
    = \int \mathcal{N}(y|\gamma_{yz}z+\gamma_{yt}t,s_y^2)\mathcal{N}(z|0,1)dz\nonumber\\
    &= \mathcal{N}(y|\gamma_{yt}t,\gamma_{yz}^2+s_y^2) = \mathcal{N}(y|\frac{\textrm{Cov}(t,y)}{\textrm{Var}(t)}t,\textrm{Var}(y) - \textrm{Cov}(t,y))
\end{align}
The third equality was obtained with standard integration. This also corresponds to the conditional distribution formula for bivariate Gaussians, $p_\theta(y|do(t))=p(y|t)$. $\square$

\begin{figure}
    \centering
    \includegraphics[width=\textwidth]{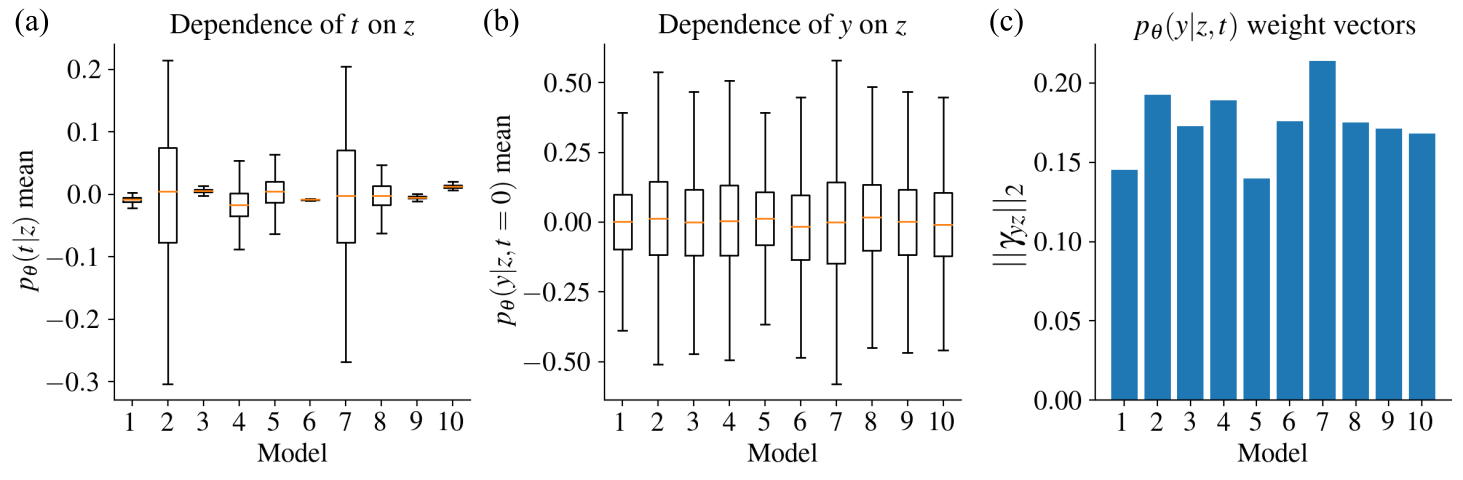}
    \caption{Visualizations of the latent space use of the ten models trained with proxy loss scaling equal to zero for the data with repeated proxies. (a) Means of $p_\theta(t|z)$ distributions with 10000 samples of $z$ from the prior of CEVAE. Aside from models 2 and 7, the mean is almost not dependent on $z$ at all. (b) Corresponding means of $p_\theta(y|z,t)$ distributions with 10000 samples of $z$ with $t=0$. Clearly, $y$ is somewhat dependent on $z$ in all of the models. (c) The $L_2$ norms of the $z$ weight vectors of the $p_\theta(y|z,t)$ predictors. (The $p_\theta(y|z,t)$ predictors were linear and the latent spaces were 10-dimensional in this experiment.) Although $y$ is dependent on $z$ in all of the models, the $z$ weight vectors $\gamma_{yz}$ are much smaller than $\gamma_{yt}$, which was around 1.24 for all models. }
    \label{fig:scaling_zero_z_dependencies}
\end{figure}

Other solutions to the group of equations are possible in principle, but in practice, the training usually converges to a solution similar to these ones, as witnessed in the repeated proxy experiment when loss scaling was set to zero. To take a closer look at these solutions, Fig.~\ref{fig:scaling_zero_z_dependencies} visualizes the dependence of $y$, $t$, and $z$ for the trained models. For all of the models, $y$ is somewhat dependent on the $z$ (nonzero values of $\gamma_{yz}$), although not as much as it is on $t$. The treatment $t$, on the contrary, is almost not dependent at all on $z$, implying that the models correspond to solution group 2. In models 2 and 7, however, $z$ does affect the treatment $t$ a small amount as well, and these probably don't match that well with the analytical solutions explained above.

Note that while the proof applies strictly speaking only to the 1D linear CEVAE, the result is true for a neural-network parameterized CEVAE as well if it is able to represent the same conditional distributions. The solution is a global optimum with both parameterizations since already $p_\theta(t,y)=p(t,y)$, and thus it's not possible to improve the ELBO according to Eq.\ref{eq:asym_marg_likelihood}.

\FloatBarrier

\section{Experiment details}

\subsection{Computing equipment, time taken to run experiments and code}

The experiments were performed with two computers: A desktop computer containing an Intel i5-6500 processor and an Nvidia GTX 970 graphics card, and a laptop containing an AMD A12-9720P processor. Most of the experiments \ref{sec:lingandetails}-\ref{sec:twinsdetails} in this section took at most a single night to run with the computing equipment, although the binary data experiment took approximately an entire day. The basic Linear-Gaussian and binary experiments were conducted with the laptop, while the others were done with the desktop computer. The graphics card was used for the proxy MNIST data set, while the processor of the desktop was used for the rest.

The code for running the experiments is provided in the following github page: \url{https://github.com/severi-rissanen/critical_look_causal_dlvms}. 

\subsection{Linear-Gaussian data}\label{sec:lingandetails}
\textbf{Data generating parameters} The generating parameters were sampled with the following process: First, all of the standard deviations $\sigma$ were generated from a Gamma(1,5) distribution. Then, the structural coefficients $c$ were got by first sampling from a Gamma(0.3,4) distribution, multiplying with the corresponding $\sigma$ and adding the result to $\sigma/2$, and uniformly randomly flipping to a negative value. This resulted in the ratio $\frac{c}{\sigma}$ being not too close to zero while keeping the absolute values of $\sigma$ and $c$ roughly in the scale of 1. Too low a ratio for proxy, for instance, would mean that the proxy would be effectively very uninformative, and could cause even the analytical methods to fail. The generated parameters for the main experiment were $c_1 = 1.03$, $c_2 = 1.47$, $c_{yz} = 0.71$, $c_{yt} = -0.62$, $\sigma_{x_1} = 0.65$, $\sigma_{x_2} = 0.96$, $\sigma_{t} = 1.25$ and $\sigma_y = 0.48$. 

\textbf{Estimation models} The default setup for the full, neural network parameterized CEVAE was so that each conditional distribution was represented with a three-layer MLP with a layer width of 30, using ELU activations. The (standard) assumption in the parameterization was that the outputs are normally distributed with a diagonal covariance for each network in the encoder and decoder. Thus, the final layers had twice the amount of heads than the output dimension, one for each mean and one for each standard deviation. There were four networks: The encoder, the proxy generation network ($p_\theta(x|z)$), the $t$ generation network ($p_\theta(t|z)$) and the $y$ generation network ($p_\theta(y|z,t)$). The dimension of the latent space was 10 for the default model. In the linear versions of CEVAE, the conditional distributions were defined with simple linear layers, and the standard deviations were separate, shared parameters used for all inputs. 

\textbf{Training} The Adam optimizer was used for all models in this work. The neural network-based models were trained with 300 epochs, as that provided good loss function convergence for all data sizes. For the linear models, we used 500 epochs. The learning rates were annealed exponentially from 0.01 to 0.001. The batch size was 200. 10 data sets were sampled for each data size, and the models were trained once for each data set. The results from training to them provide the box plots in the results. 

\begin{figure}
    \centering
    \includegraphics[width=\textwidth]{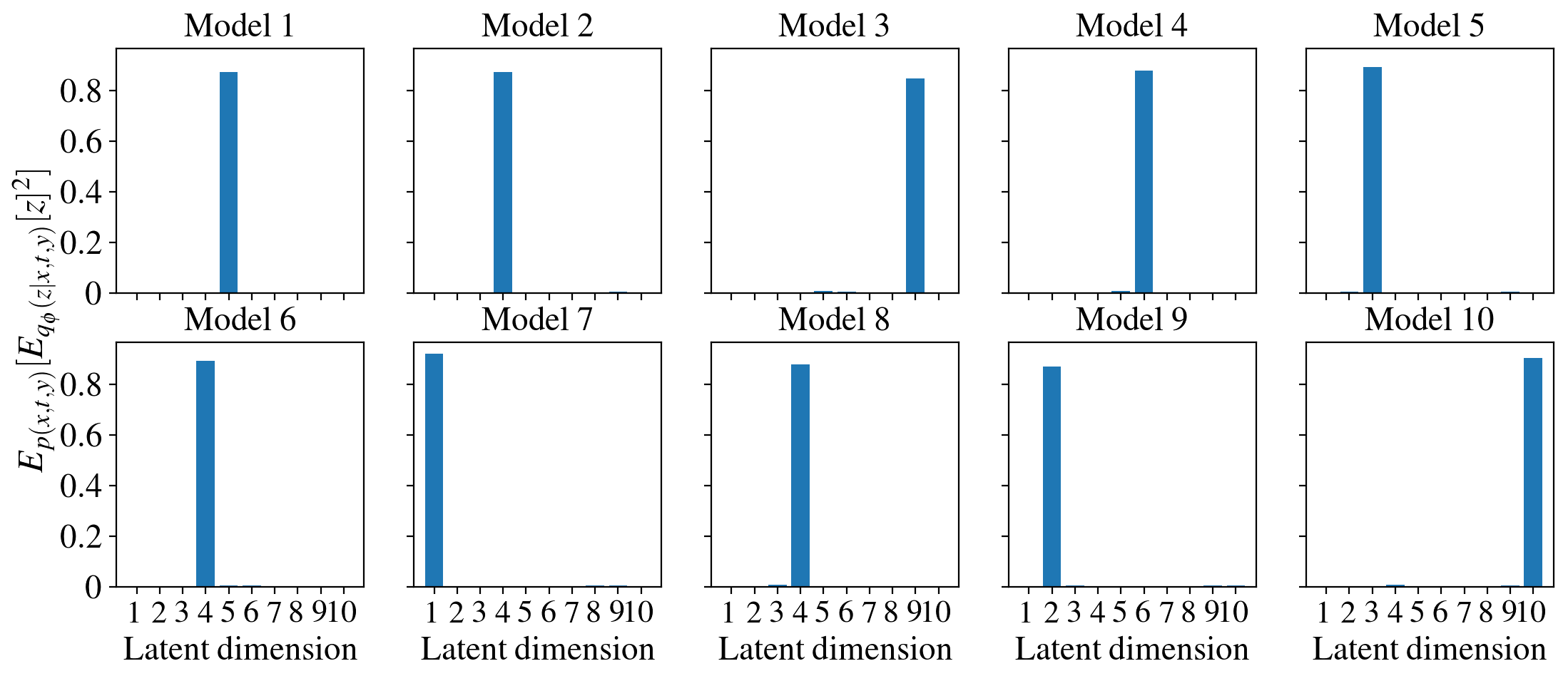}
    \caption{The squared expected values of the posterior approximations for the full 10D CEVAE with linear-Gaussian data, averaged over their respective data sets. Each of the models was trained on one of the linear-Gaussian data sets of size 20000.}
    \label{fig:posteriorcollapse}
\end{figure}

\textbf{Posterior collapse} Figure \ref{fig:posteriorcollapse} visualizes the posterior collapse phenomenon for the 10 models trained with a data size of 20000. It shows the squared expected values $\mathbb{E}[q_\phi(z|x,t,y)]^2$ for each of the 10 dimensions, averaged over the data set (in other words, the variances of the encoder means for the data set). We see that all of the models use only one of the latent dimensions, while all the unused dimensions in the posterior approximation fall back to the mean of the prior. 

\textbf{Failed 2D estimation} In the linear, 2D CEVAE experiment with failed estimation, we tried to initialize the model so that the parameters included aspects of the "correct" parameters, but were also sufficiently different to lead the model to incorrect estimation after training. The chosen initialization was also not itself a minimum of the loss function. Figure \ref{fig:lingan2d_convergence}a shows the initialization. We tried to be very careful with the training by increasing the batch size to 1000, setting the learning rate to 0.001 and training until the model appeared to converge. In Fig.\ref{fig:lingan2d_convergence}b-c we plot the losses and estimates for the $c_{yt}$ coefficients as the training progressed. While the custom initialization results in an indistinguishable loss, the resulting causal effect is clearly wrong. 

\textbf{Experiment with 10D linear CEVAE with an attempt to avoid posterior collapse} To take a deeper dive into the effect of posterior collapse on causal effect estimation with CEVAE, we designed an experiment with the purpose of maximizing disentanglement in the latent space of CEVAE. Before, a 2D linear CEVAE was somewhat prone to not posterior collapse completely, so we tried increasing the latent dimensionality to ten. We also annealed the KL divergence term from a low value to the regular one during training to promote disentanglement. 

\begin{figure}
    \centering
    \includegraphics[width=\textwidth]{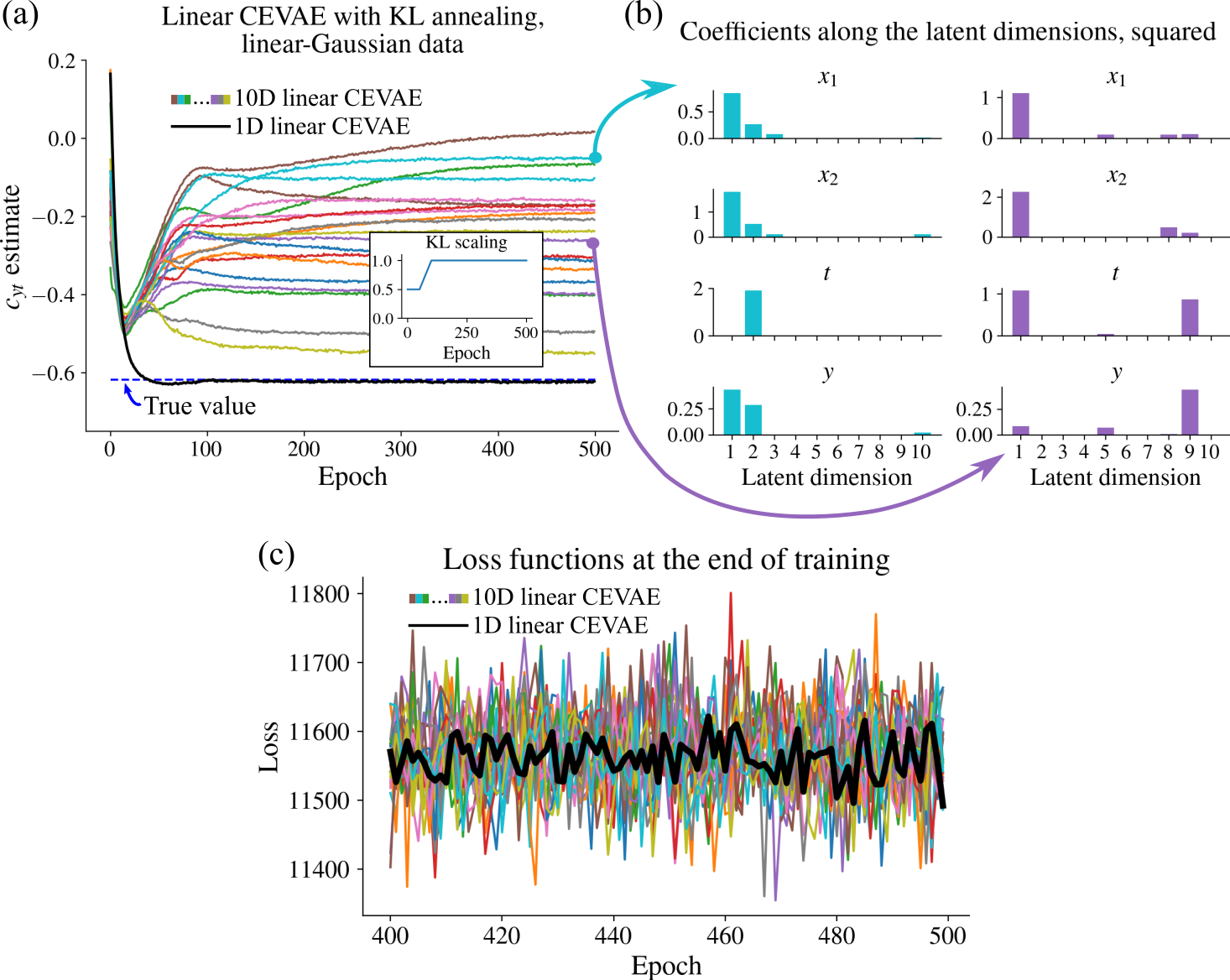}
    \caption{Results of the experiment where we try to discourage posterior collapse by annealing the KL divergence term from a low value at the beginning of training. All runs use a linear-Gaussian data set of size 2000. (a) $c_{yt}$ estimates for a linear CEVAE with a ten-dimensional latent space, trained twenty times, and for a linear CEVAE with a one-dimensional latent space. The 10D models do not estimate the causal effect correctly. (b) Coefficients of different linear predictors in the decoders of two selected models. Both use more than one dimension. (c) Loss functions towards the end of training. The 1D linear CEVAE doesn't reach a lower loss than the models that failed at estimation.}
    \label{fig:linear10D_postcollapse_fail}
\end{figure}

\begin{figure}
    \centering
    \includegraphics[width=\textwidth]{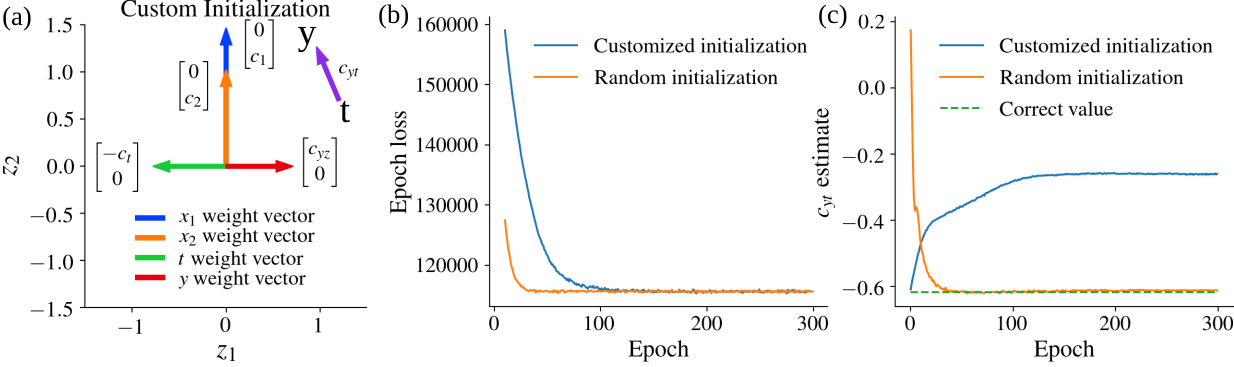}
    \caption{(a) Custom initialization of the 2D linear CEVAE that results in the wrong causal effect. The variances were set to the correct values and the posterior approximation was set so that it was the true posterior given the initialized decoder parameters. (b) Epoch losses of the custom initialized CEVAE and a successful, randomly initialized 2D linear CEVAE for comparison. (b) Estimates of $c_{yt}$, taken as the coefficient of $t$ in the linear predictor for $y|z,t$ in the decoder. }
    \label{fig:lingan2d_convergence}
\end{figure}

Figure \ref{fig:linear10D_postcollapse_fail}a shows the results for a sample of of size 2000. The subpanel shows the scaling of the KL divergence term. We trained a 10D linear CEVAE twenty times, and a 1D linear CEVAE for comparison. The ten-dimensional model doesn't estimate the causal effect correctly, and instead, the estimates converge towards random values. The one-dimensional model, on the other hand, works correctly and settles on the correct value soon after the KL divergence term is returned to normal. From panel b we see that the ten-dimensional models indeed do use more than one dimension in reconstruction, similar to the situation with the failed 2D linear CEVAE in the main text. Importantly, the loss function at the end of training for the correct, one-dimensional model is indistinguishable from the ones of failed, higher-dimensional models. Thus, it's plausible that all of the models ended up in a global minimum or in a state very close to being one, and the model is unidentifiable with respect to causal effect estimates. 

\subsection{Binary data}\label{sec:binarydetails}

\textbf{Data generating parameters} The data generating parameters were generated from a Dirichlet(2) distribution for the prior $p(z)$ as well as all the conditional distributions. The distribution was chosen so that probabilities close to 0 or 1 would be unlikely, as too high or low probabilities can result in a pathological case where the data is very difficult to estimate from. The parameters for the main experiment were $p(z=1) = 0.56$, $p(x_1=1|z=0) = 0.56$, $p(x_1=1||z=1) = 0.73$, $p(x_2=1|z=0) = 0.94$, $p(x_2=1|z=1) = 0.26$, $p(t=1|z=0) = 0.71$, $p(t=1|z=1) = 0.16$, $p(y=1|z=0,t=0) = 0.57$, $p(y=1|z=0,t=1) = 0.36$, $p(y=1|z=1,t=0) = 0.17$ and $p(y=1|z=1,t=1) = 0.04$. 

\textbf{Estimation models} The default setup for the full CEVAE was the same as in the Linear-Gaussian experiment, except that the neural network heads for the decoder were transformed trough logistic link functions to probabilities, and the output was interpreted as a standard Bernoulli distribution. We also tried a model with a binary latent space, which was otherwise the same as the regular model, except the encoder had just one head, and the output was similarly interpreted as a Bernoulli distributed variable. The probability parameter of the Bernoulli distributed prior $p(z)$ was included as a learnable parameter. The expectation $\mathbb{E}_{q_\phi(z_i|x_i,t_i,y_i)}[\log p_\theta(x_i,t_i,y_i|z)]$ was calculated directly by passing both values of $z$ through the decoder weighting the log probabilities with $q_\phi(z_i|x_i,t_i,y_i)$. 

\textbf{Training} The models were trained for 300 epochs with the Adam optimizer, with learning rate annealing from 0.01 to 0.0005, as that seemed to produce good loss function convergence. 10 data sets were sampled for each data size, and the batch size was 200. 

\textbf{Binary CEVAE results}

\begin{figure}
    \centering
    \includegraphics[width=\textwidth]{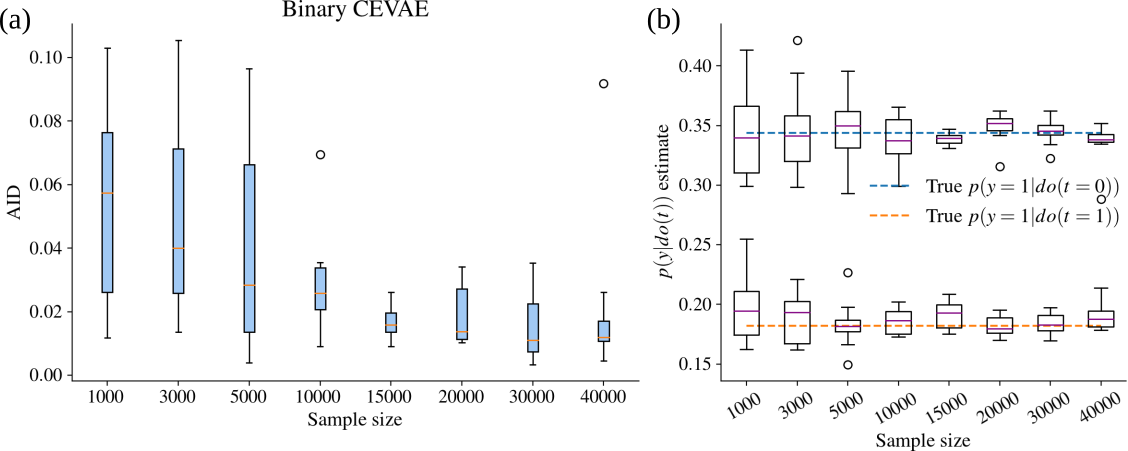}
    \caption{The AID and $p(y|do(t))$ estimates as a function of sample size for the binary CEVAE and binary data. }
    \label{fig:binary_cevae_result}
\end{figure}

Figure \ref{fig:binary_cevae_result} shows the results from the binary CEVAE. It performs very well, and the causal effect estimates become better as the sample size is increased. However, the model doesn't work as well for all data generating processes, such as some of the ones in Sec.~\ref{sec:binaryreplication}, where it seemed that the model can get stuck in local minima easily.

\subsection{Linear-Gaussian data with redundant noise}

\textbf{Data generation} The parameters for the first phase of the data generating distribution were $c_1 = 1$, $c_2 = 2$, $c_t = 0.5$, $c_{yz} = 0.6$, $c_{yt} = 1$, $\sigma_1 = 0.5, \sigma_2 = 0.7$, $\sigma_3 = 20$, $\sigma_t = 1$ and $\sigma_y = 1$ ($c_3$ was 0). After generating data from the linear-Gaussian distribution defined with these parameters, an additional orthogonal transformation was applied on the proxies so that the matrix was
\begin{equation}
    R = \begin{bmatrix} \cos\alpha & -\sin\alpha & 0\\\sin\alpha & \cos\alpha & 0 \\ 0 & 0 & 1 \end{bmatrix} \begin{bmatrix} \cos\beta & 0 & \sin\beta \\ 0 & 1 & 0 \\ -\sin\beta & 0 & \cos\beta \end{bmatrix} \begin{bmatrix} 1 & 0 & 0 \\ 0 & \cos\gamma & -\sin\gamma \\ 0 & \sin\gamma & \cos\gamma \end{bmatrix}
\end{equation}
where $\alpha=\frac{\pi}{4}$, $\beta=\frac{\pi}{4}$ and $\gamma=\frac{\pi}{4}$. Here $\alpha, \beta$ and $\gamma$ are also called the yaw, pitch and roll angles in the 3D space. 

\textbf{Estimation models and training} The parameters of CEVAE and training were otherwise identical to the full CEVAE linear-Gaussian experiment, but the latent space was changed to one-dimensional and two-dimensional for the two setups explained in the main text. 

\subsection{Linear-Gaussian data with copied proxies}

\textbf{Data generation} The data generating parameters were $c_1 = 1$, $c_2 = 1$, $c_t = 0.5$, $c_{yz} = 0.6$, $c_{yt} = 1$, $\sigma_1 = 2$, $\sigma_2 = 2$, $\sigma_t = 1$ and $\sigma_y = 1$. $\tilde x_1$ and $\tilde x_2$ were copies of $x_1$ and $x_2$, with Gaussian noise of standard deviation 0.1 added.

\textbf{Estimation model} Other experimental parameters were exactly the same as for the full CEVAE in the linear-Gaussian experiment, except that we used a linear predictor for $y$ to make the results easier to interpret. 

\textbf{Training} Training took only 100 epochs when trying out different loss scaling values, since that was clearly enough for convergence. The sample size was 20000 for the loss scaling experiment.

\subsection{Proxy IHDP data}

\textbf{Consent and personally identifiable info.} The IHDP data is for public use and the data is not personally identifiable, as far as we are aware. 

\begin{figure}[h]
    \centering
    \includegraphics[width=\textwidth]{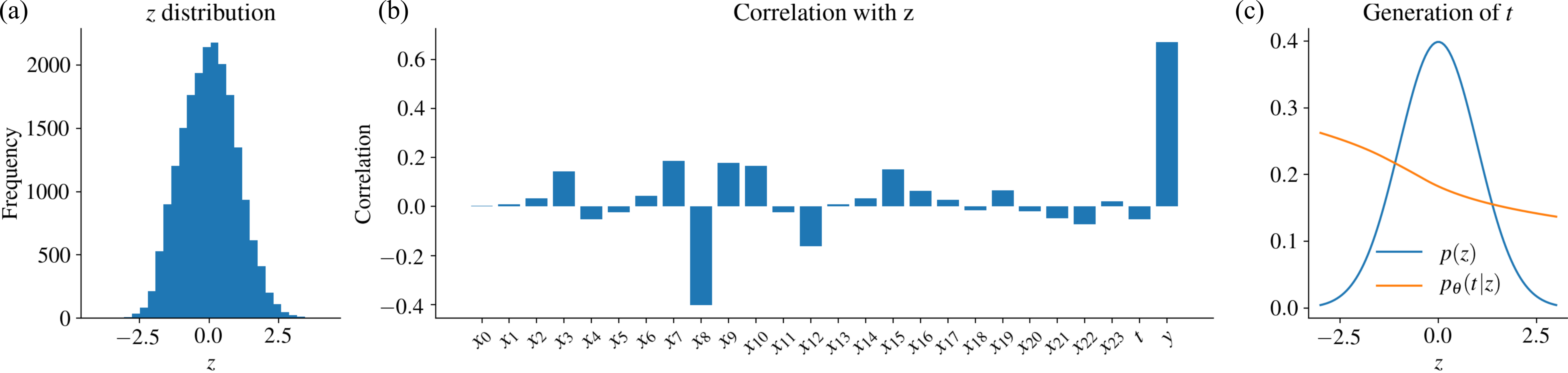}
    \caption{Properties of the IHDP data generating process. (a) The distribution of the chosen unobserved confounder. (b) Correlation of the chosen $z$ with the proxies, $t$ and $y$. (c) Visualization of the $p(t|z)$ function in the data generating process.}
    \label{fig:ihdp_datageneration}
\end{figure}

\textbf{Data generation} The data generating VAE was defined so that the encoder and decoder were MLPs with 3 layers and layer width 30 and a latent dimension of 10. The decoder was structured with the assumption that continuous variables had a Gaussian distribution at the outputs of the neural networks and categorical variables were categorically distributed with a softmax layer or logistic link function at the end for binary variables. The fourth variable in the IHDP data set was transformed to categorical since only four distinct values occured in the data set. Variances were estimated individually for each sample for the continuous variables with the neural networks. We chose the fifth variable as the unobserved confounder, and left the rest as the proxies. The treatment values were generated by first fitting a feed-forward neural network from the hidden confounder to the treatment values in the original data set and then sampling the treatments based on the probability given by the neural network. The $y$ values were then generated by $y^i = z^i + t^i\cdot\textrm{ATE} + \epsilon^i$, where $\epsilon^i$ is Gaussian noise with a standard deviation of 1 and ATE is the average treatment effect, which was also set to 1. ATE was set to 1 and the added Gaussian noise on $y$ had a standard deviation of 1. Figure \ref{fig:ihdp_datageneration} presents some aspects of the generating process: The chosen unobserved confounder is approximately normally distributed and it is clearly correlated with many of the observed variables. Panel (c) also visualizes the dependence of $t$ on $z$.

\textbf{Estimation model} The CEVAE model used for the data was defined similarly as the data generating VAE, with the addition of $t$ and $y$ prediction networks in the decoder. Another difference was that there were two separate networks in the decoder for $y$ generation that were chosen based on the value of $t$ on each pass. Likewise, the encoder consisted of four parallel networks that were chosen based on the value combination of $t$ and $y$ for each sample. The aim was to force the model to take $t$ and $y$ in to account in the reconstruction process, in a similar way as in the original CEVAE publication. Also, the fifth variable in the IHDP data naturally wasn't included in the proxies. 

\textbf{Training} Training was done for 200 epochs with a batch size of 200 and exponential learning rate annealing from 0.001 to 0.00001. The sample size for the loss scaling experiment was 20000. 

\begin{figure}[h]
    \centering
    \includegraphics[width=0.7\textwidth]{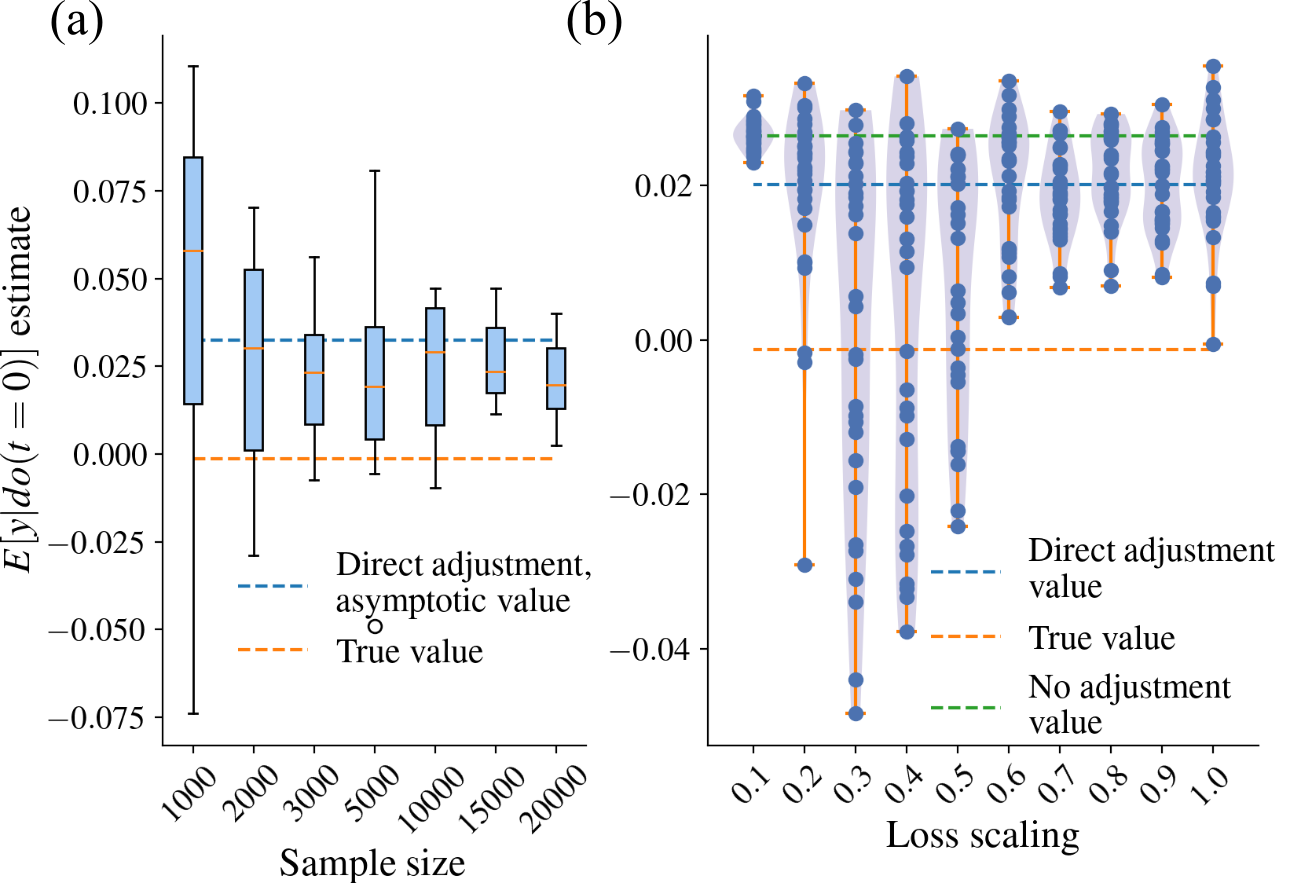}
    \caption{The $\mathbb{E}[y|do(t=0)]$ estimates with respect to sample size and loss scaling with sample size set to 20000. }
    \label{fig:ihdp_pydot0}
\end{figure}

\textbf{Additional results} Figure \ref{fig:ihdp_pydot0} shows the experimental results for $\mathbb{E}[y|do(t=0)]$. We see that the conclusions are the same as for $\mathbb{E}[y|do(t=1)]$, which was presented in the main text.

\subsection{Proxy MNIST data}\label{sec:proxymnistdetails}

\textbf{Consent and personally identifiable info.} The MNIST data is open access and the data is not personally identifiable, as far as we are aware. 

\textbf{Data generation} The data generating GAN had a simple four-layer convolutional architecture in the discriminator and similarly four transpose convolutional layers in the generator. The exact details are in the code accompanying the article. The $t$ and $y$ values were Bernoulli distributed with a probability dictated by the value of the unobserved confounder (the first latent dimension of the GAN) through a logistic link function:
\begin{align}
    t|z&\sim\textrm{Bern}(\sigma(a_t z + b_t))\nonumber\\
    y|z,t&\sim\textrm{Bern}(\sigma(a_{y,1} z + b_{y,1})t + \sigma(a_{y,0} z + b_{y,0})(1-t)) \nonumber
\end{align}
where $a_t=1$, $b_t=0.5$, $a_{y,1}=2$, $b_{y,1}=0.5$, $a_{y,0}=2$ and $b_{y,0}=-0.5$. The additional linear-Gaussian proxy was generated from the unobserved confounder using $c=1$ and $\sigma=1$.

\textbf{Estimation model} The CEVAE model had was structured in a similar way as the GAN, with four transpose convolutional layers in the image part of the decoder. The other neural networks for the extra linear-Gaussian proxy, $t$ and $y$ were three-layer MLPs of width 30. Two NNs were defined for $y$, chosen depending on the value of $t$, the attempt being to define the conditional distributions as well as possible for the task. The encoder consisted of three parts. First, we had the four transpose convolutional layers starting from the images and ending up with 40 outputs. Second, we had a set of four fully connected linear layers with the 40 previous outputs and the additional proxy variable as inputs, with an output size of 125. One of the four networks was then chosen on each pass based on the value combination of $t$ and $y$. The third part was similarly a set of 4 fully connected linear layers with input size 25 and output size 40, where one of the networks was again chosen based on the values of $t$ and $y$. Again, the aim was to force the encoder to take $t$ and $y$ in to account. The final output was used to define the means and variances of the variational approximation in the 20-dimensional latent space. The convolutional and transpose convolutional layers were different from the corresponding GAN layers in that some of the kernel sizes, strides and paddings were changed around just so that the estimation model didn't match exactly with the data generation model. 

\textbf{Training} The models were trained for 500 epochs (as shown in the Figures in the main text) with a batch size of 1000 and exponential schedule learning rate annealing from 0.003 to 0.001. 

\begin{figure}
    \centering
    \includegraphics[width=0.4\textwidth]{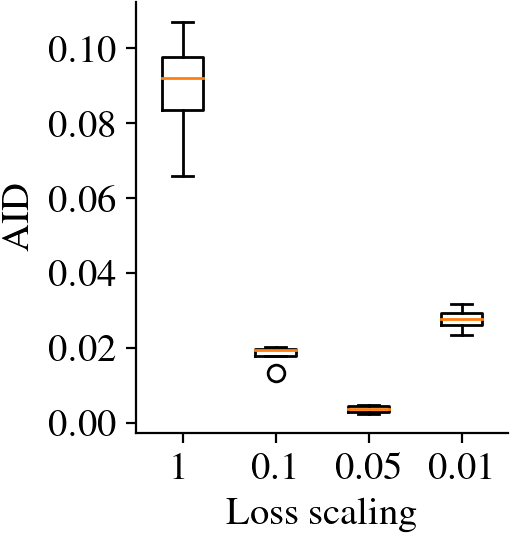}
    \caption{The AID values got by running CEVAE multiple times for each image reconstruction loss scaling factor. The data set was the same data set of size 10000 that was used to produce the figure in the main text. }
    \label{fig:mnist_AID}
\end{figure}

\textbf{Repeated experiment} Figure \ref{fig:mnist_AID} shows the AID values of the same experiment run multiple times for each image reconstruction loss scaling value. The results confirm the pattern noted in the main text: The causal estimates are far off with scaling equal to 1, but get closer as we reduce it to 0.1 and the AID approaches almost zero with scaling 0.05. Decreasing further, the AID starts increasing again, indicating that there is an optimal value between zero and one. 

\subsection{Twins data}\label{sec:twinsdetails}

The original CEVAE paper presented the Twins data as an example of a data set where the causal effects are identifiable in principle (with knowledge of the data generating process), but the distribution of $z, t$, and $y$ is derived from real data. They reported that the model produced better estimates than a direct adjustment to the proxies, but our focus is on the asymptotic performance and the identifiability of the model. It is interesting from our point of view also because the unobserved confounder is categorical, and it provides us with a further example of a misspecified latent variable in addition to the binary data experiment. 

\textbf{The data.} The data consists of information about twin births and infant mortalities in the USA during the years 1989-1991. The "treatment" is defined to be the variable that tells us whether the particular newborn was the heavier or lighter of the twins ($t=0$ for the lighter twin, $t=1$ for the heavier). The outcome $y$ is a binary variable that indicates mortality within a year after birth. There are 46 covariates relating to the parents, out of which the GESTAT10 feature is chosen as the unobserved confounder. It consists of ten categories indicating the number of gestation weeks before birth. An observational study is then simulated by choosing one of the $t$ values for each twin pair based on the value of the confounder: $\textrm{Bern}(\sigma(w_o^T\vec v + w_h(z/10 - 0.1)))$, where $\vec v$ is a vector that contains the other covariates. Having the other covariates in the formula produces some additional randomness to the assignment of $t$. The weight vectors $w_o$ and $w_h$ are sampled from $w_o\sim\mathcal{N}(0,0.1\cdot I), w_h\sim\mathcal{N}(5,0.1)$. Proxies are created based on the categorical confounder by taking a one-hot representation, copying it three times, and randomly flipping each binary feature with some probability. The original paper used probabilities ranging from 0.05 to 0.5, and we use 0.2 in our experiments. We generated $w_o$ and $w_h$ once and used those in the experiments to reduce the amount of randomness. The used $w_h$ weight was 5.030. Table \ref{tab:twins_wo} lists the coefficients of the weight vector $w_h$.

\begin{table}[]
    \centering
    \begin{tabular}{cccccccccc}
         $w_{o,0}$&$w_{o,1}$&$w_{o,2}$&$w_{o,3}$&$w_{o,4}$&$w_{o,5}$&$w_{o,6}$&$w_{o,7}$&$w_{o,8}$&$w_{o,9}$\\ 
0.068&0.167&-0.187&0.150&-0.177&-0.101&0.143&0.047&0.112&-0.039\\ 
\hline
$w_{o,10}$&$w_{o,11}$&$w_{o,12}$&$w_{o,13}$&$w_{o,14}$&$w_{o,15}$&$w_{o,16}$&$w_{o,17}$&$w_{o,18}$&$w_{o,19}$\\ 
-0.087&-0.098&-0.077&-0.070&-0.069&-0.063&-0.156&-0.070&-0.037&0.143\\ 
\hline 
$w_{o,20}$&$w_{o,21}$&$w_{o,22}$&$w_{o,23}$&$w_{o,24}$&$w_{o,25}$&$w_{o,26}$&$w_{o,27}$&$w_{o,28}$&$w_{o,29}$\\ 
0.141&-0.035&0.204&0.153&-0.070&0.273&-0.008&-0.143&0.109&-0.155\\ 
\hline 
$w_{o,30}$&$w_{o,31}$&$w_{o,32}$&$w_{o,33}$&$w_{o,34}$&$w_{o,35}$&$w_{o,36}$&$w_{o,37}$&$w_{o,38}$&$w_{o,39}$\\ 
0.127&-0.017&0.005&0.077&0.081&0.019&-0.026&-0.078&-0.172&-0.051\\ 
\hline 
$w_{o,40}$&$w_{o,41}$&$w_{o,42}$&$w_{o,43}$&$w_{o,44}$&$w_{o,45}$\\ 
0.114&-0.134&0.018&-0.104&-0.132&0.123
    \end{tabular}
    \caption{The $w_o$ weight vector used in the Twins data experiments. It was sampled from the distribution $\mathcal{N}(0,0.1\cdot I)$.}
    \label{tab:twins_wo}
\end{table}

Because we want to consider data sets that are larger than the original data set, we decided to use bootstrapping to generate new sets that are distributed according to the empirical distribution of the Twins data. So the full data generating process was that $z$, $v$, and both $y$ values were sampled with replacement for a given number of times from the initial data, after which the $t$ values and the proxies were sampled with the procedure explained earlier. The mortality of the lighter twin is $p(y=1|do(t=0))=18,9\%$ whereas the heavy twin mortality is slightly lower, $p(y=1|do(t=1)))=16,4\%$.

\textbf{Consent and personally identifiable info.} The NCHS data the Twins data set is based on is for public use and the data is not personally identifiable, as far as we are aware.

\textbf{Estimation model.} The used CEVAE model was the regular full CEVAE, with the exception that layer width was 50 instead of 30 in the encoder. We made this choice because the number of observed variables was over 30, and using the regular width would have acted as a bottleneck and could have caused sub-optimal results.

\textbf{Training.} In the experiment with varying sample sizes and multiple different data sets, the models were trained for 1000 epochs with a batch size of 500 and learning rate annealing going from 0.004 to 0.0002. With the experiments where the model was trained multiple times on one data set, the number of epochs was 2000 to make sure that the estimates seemed to converge to some value. 

\begin{figure}
    \centering
    \includegraphics[width=\textwidth]{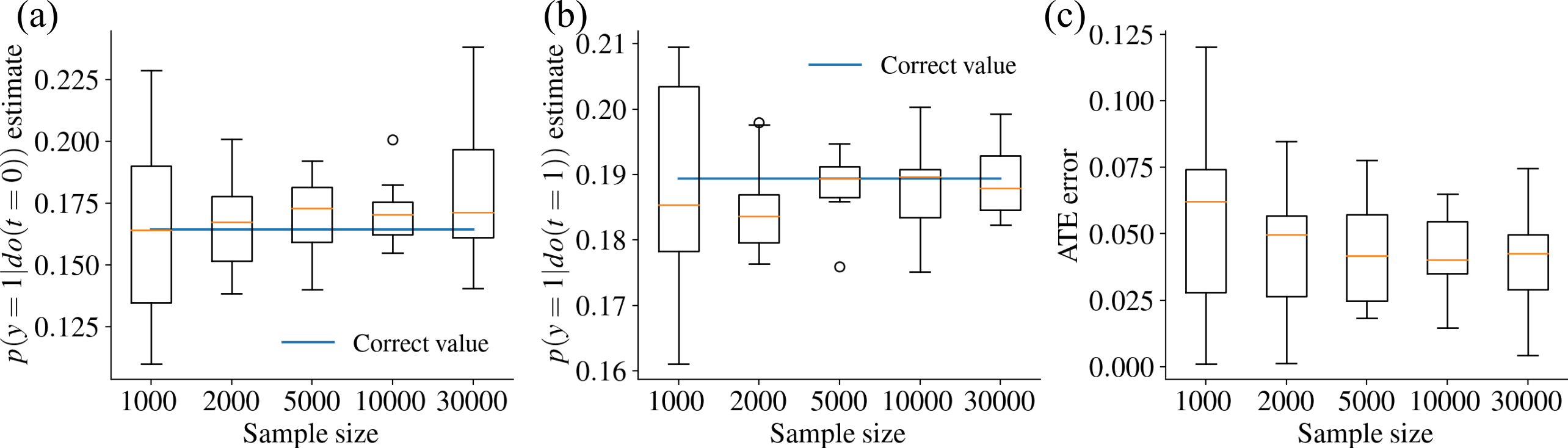}
    \caption{Causal effect estimates and the ATE errors for the Twins data set.}
    \label{fig:twins_1000to50000}
\end{figure}

\begin{figure}
    \centering
    \includegraphics[width=\textwidth]{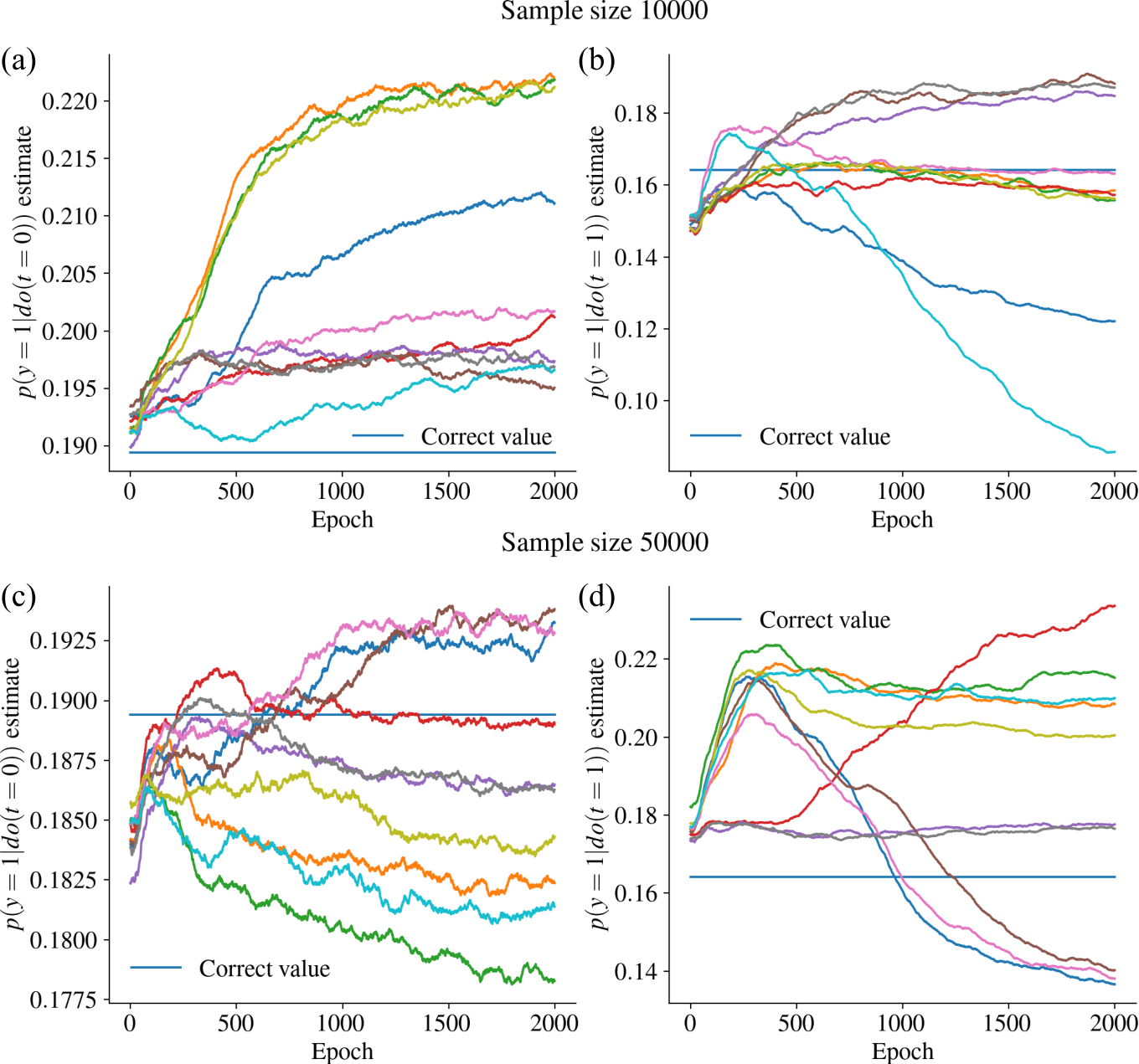}
    \caption{Causal effect estimates with respect to the number of epochs during training for the Twins data. The estimates are averaged over a window of 100 epochs to smooth out temporary variation. (a),(b) Ten runs for one data set of size 10000. (c),(d) Ten runs for one data set of size 50000.}
    \label{fig:twins_repeated_estimate}
\end{figure}

\begin{figure}
    \centering
    \includegraphics[width=0.8\textwidth]{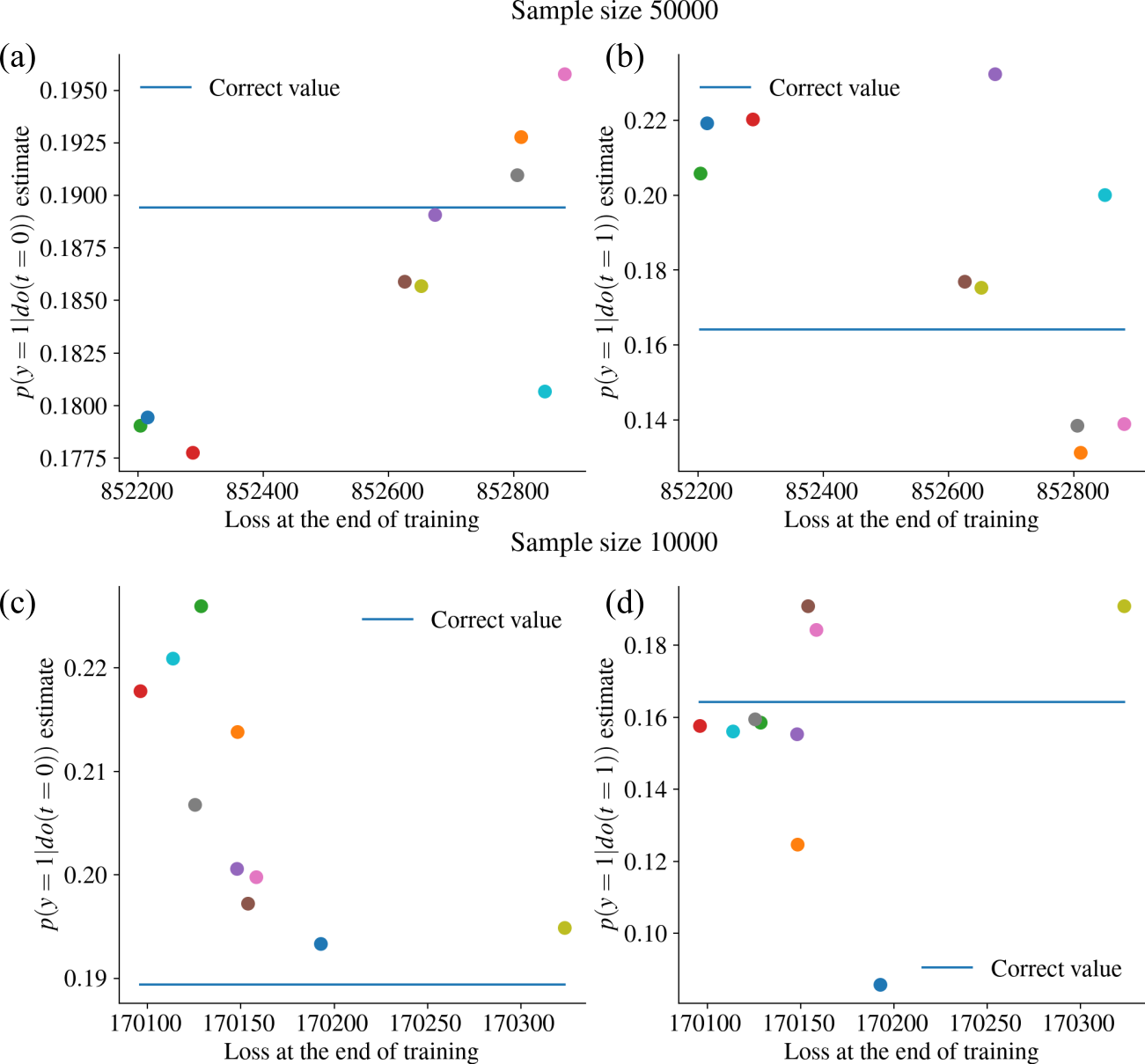}
    \caption{The causal effect estimates with respect to the loss at the end of training for the Twins data set. The losses and estimates are averages over the final one hundred epochs. (a),(b) Estimates for a data set of size 10000. (c),(d) Estimates for a data set of size 5000.}
    \label{fig:twins_repeated_estimate_loss}
\end{figure}

\textbf{Results. } Figure \ref{fig:twins_1000to50000} shows the causal effect estimates and the ATE error as a function of sample size. While the ATE errors are roughly in the same order of magnitude as the ones reported in the original paper for proxy noise level 0.2, it's not clear that the estimates converge towards the correct values. In fact, it looks like they are not converging towards any value in particular, even though quite a lot of effort was put into careful training. To take a closer look at this issue, we trained the model multiple times for single generated data sets of sizes 10000 and 50000 and recorded the causal effect estimates as the training progressed. Figure~\ref{fig:twins_repeated_estimate} shows the results for both data sets. With a sample size of 10000, the model isn't able to get a consistent estimate, and increasing the amount of data doesn't seem to help either, since the same happens with a sample size of 50000. It seems that the model ends up in different kinds of minima that produce different kinds of estimates. Figure \ref{fig:twins_repeated_estimate_loss} shows that using the loss value at the end of training can not be used to choose a well-performing model, as models that got (slightly) smaller losses than others produce even worse estimates than many models with a higher loss. The losses are also very close to each other, so one could say that this is a case of "effective" model unidentifiability, where it's not possible to find a single estimate.

\FloatBarrier

\section{Replicated experiments}

\subsection{Linear-Gaussian data}

\begin{figure}[h]
    \centering
    \colorbox{white}{\includegraphics[width=0.8\textwidth]{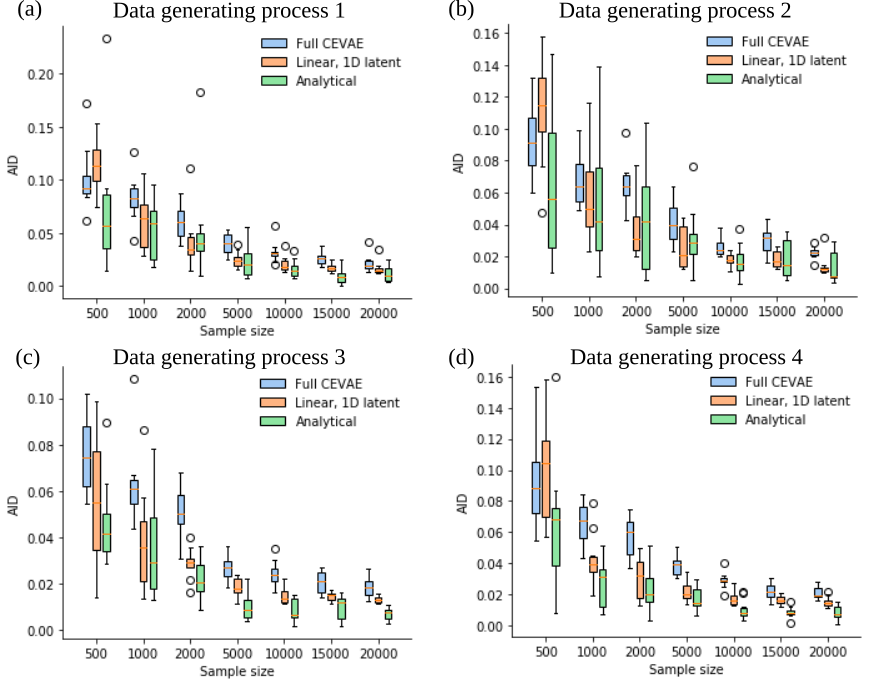}}
    \caption{The AID values with respect to sample size for the replicated linear-Gaussian experiments. }
    \label{fig:lingan_repeat}
\end{figure}

We sampled four other data-generating distributions with the process detailed in Section \ref{sec:lingandetails} and ran the same experiment comparing the convergence of AID values on each process. The parameters are presented in Table \ref{tab:replicated_lingan_parameters}. Figure \ref{fig:lingan_repeat} shows the results. While the convergence of AID values to zero happens differently for different data generating processes, the main conclusion stays the same. The estimates of the full CEVAE model are usually slightly less accurate than the 1D linear CEVAE and the analytical method, but they do steadily improve as the sample size increases. 

\begin{table}[h]
    \centering
    \caption{Data generating parameters for the repeated linear-Gaussian experiments.}
    \begin{tabular}{c c c c c c c c c c c}
         & $c_1$ & $c_2$ & $c_t$ & $c_{yz}$ & $c_{yt}$ & $\sigma_1$ & $\sigma_2$ & $\sigma_t$ & $\sigma_y$\\\hline
         Process 1& -0.53 & 0.92 & 0.99 & -1.15 & 0.46 & 0.71 & 1.02 & 1.14 & 0.84 \\
         Process 2& 1.05 & -0.57 & -0.83 & 0.76 & -1.38 & 1.04 & 0.77 & 0.68 & 1.11\\
         Process 3& 1.30 & -1.02 & 0.80 & 1.17 & 1.11 & 1.02 & 0.91 & 1.27 & 0.88\\
         Process 4& -1.58 & 0.80 & -0.82 & 0.99 & -1.13 & 1.28 & 0.87 & 1.04 & 0.77
    \end{tabular}
    \label{tab:replicated_lingan_parameters}
\end{table}

\subsection{Binary data}\label{sec:binaryreplication}

\begin{figure}
    \centering
    \includegraphics[width=\textwidth]{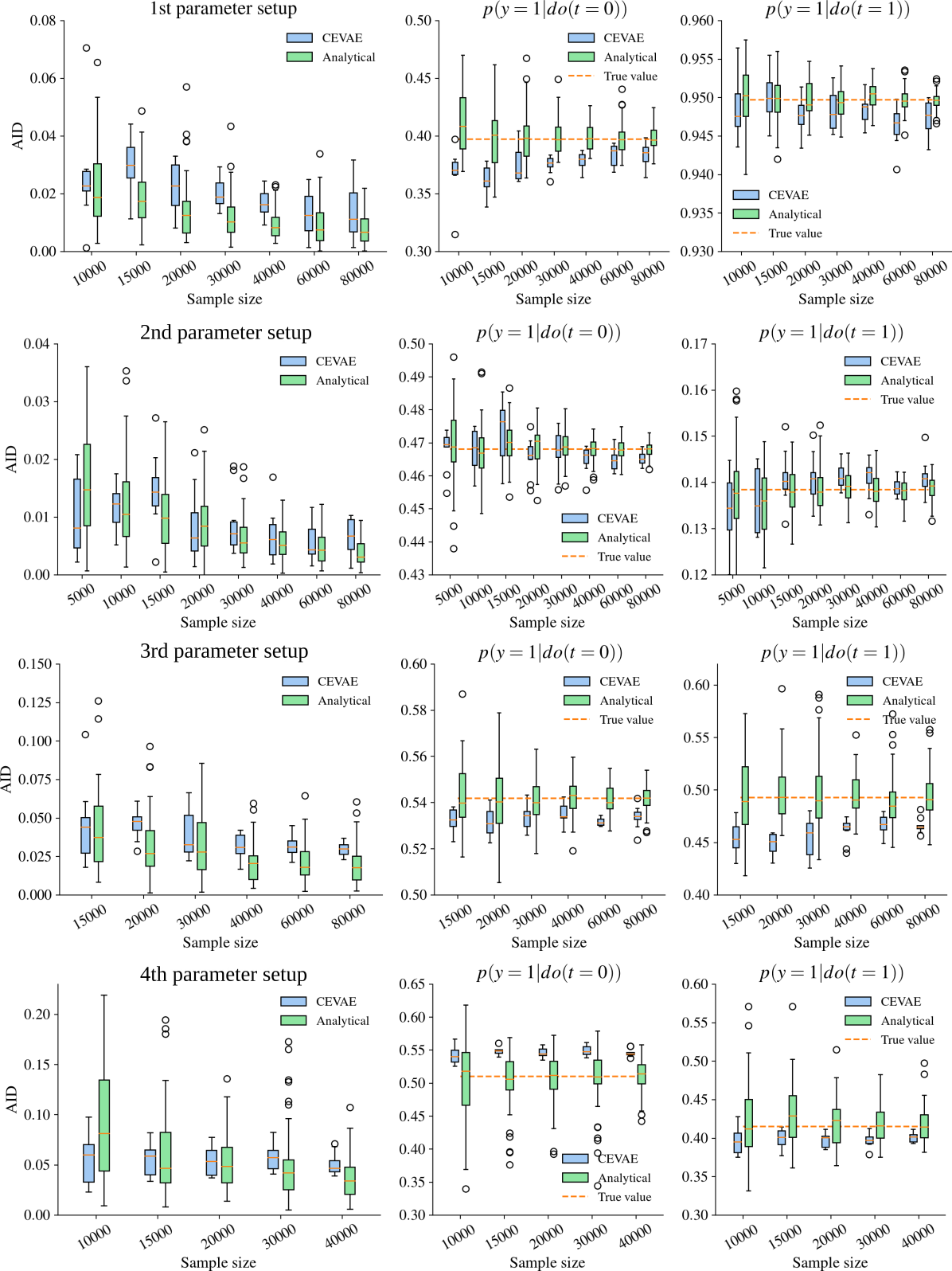}
    \caption{The results from the four additional binary data generating processes. Some outliers are not shown in order to make the plots readable.}
    \label{fig:additional_binary_experiments}
\end{figure}

Four other data generating processes were sampled from the distribution explained in Sec.\ref{sec:binarydetails}, and the same experiment as in the main text was run for each data generating process. The results are presented in Fig.\ref{fig:additional_binary_experiments}. Aside from possibly the second data set, CEVAE consistently estimates the causal effects incorrectly. The analytical method is not very accurate with small sample sizes, but the average estimate is correct, in contrast to CEVAE. The fact that CEVAE performed well especially on the second data generating process may be due to the true causal effects being very close to the effects we get without adjustment at all (compare $p(y=1|do(t=0))=0.468$ to $p(y=1|t=0)=0.473$ and $p(y=1|do(t=1))=0.138$ to $p(y=1|t=1)=0.136$).

\begin{table}[h]
    \centering
    \caption{Data generating parameters for the repeated binary data experiments.}
    \begin{tabular}{c c c c c}
         & Process 1 & Process 2 & Process 3 & Process 4\\\hline
         $p(z=1)$&0.41 & 0.49 & 0.42 & 0.45\\
         $p(x_1=1|z=0)$&0.88 & 0.24 & 0.63 & 0.30\\
         $p(x_1=1|z=1)$&0.66 & 0.73 & 0.44 & 0.44\\
         $p(x_2=1|z=0)$&0.63 & 0.53 & 0.81 & 0.33\\
         $p(x_2=1|z=1)$&0.86 & 0.63 & 0.47 & 0.65\\
         $p(t=1|z=0)$&0.51 & 0.44 & 0.19 & 0.25\\
         $p(t=1|z=1)$&0.78 & 0.29 & 0.64 & 0.78\\
         $p(y=1|z=0,t=0)$&0.21 & 0.42 & 0.49 & 0.67\\
         $p(y=1|z=0,t=1)$&0.93 & 0.12 & 0.72 & 0.54\\
         $p(y=1|z=1,t=0)$&0.66 & 0.52 & 0.61 & 0.31\\
         $p(y=1|z=1,t=1)$&0.97 & 0.15 & 0.18 & 0.26\\
    \end{tabular}
    \label{tab:replicated_binary_parameters}
\end{table}

\end{document}